\documentclass{article}

\PassOptionsToPackage{numbers, compress}{natbib}
\usepackage[main, final]{neurips_2026}

\usepackage[utf8]{inputenc} % allow utf-8 input
\usepackage[T1]{fontenc}    % use 8-bit T1 fonts
\usepackage{hyperref}       % hyperlinks
\usepackage{url}            % simple URL typesetting
\usepackage{booktabs}       % professional-quality tables
\usepackage{amsfonts}       % blackboard math symbols
\usepackage{nicefrac}       % compact symbols for 1/2, etc.
\usepackage{microtype}      % microtypography
\usepackage{xcolor}         % colors
\usepackage{amsthm}
\usepackage{xspace}

\usepackage{threeparttable}
\usepackage{multicol,multirow}
\usepackage{balance}

\usepackage{colortbl}  %
\usepackage{graphicx}
\usepackage{pifont}
\usepackage{amssymb}

\usepackage{etoc}       % \etocdepthtag.toc, \etocsettagdepth
\etocdepthtag.toc{mtchapter}
\etocsettagdepth{mtchapter}{subsubsection}
\etocsettagdepth{mtappendix}{none}

\usepackage{tcolorbox}  % \newtcolorbox
\usepackage{tikz}

\newtheorem{proposition}{Proposition}

\usepackage{amsmath}

  \DeclareMathAlphabet\mathbfcal{OMS}{cmsy}{b}{n}

  \def\0{{\bf 0}}
  \def\1{{\bf 1}}

\def\eg{\emph{e.g.}} 
\def\ie{\emph{i.e.}} 
\def\cf{\emph{c.f.}} 
\def\etc{\emph{etc.}}

\usepackage{enumitem}
\usepackage{fontawesome5}
\usepackage[ruled,vlined,linesnumbered]{algorithm2e}
\DontPrintSemicolon
\DeclareSymbolFont{txsymbolsC}{U}{txsyc}{m}{n}
\DeclareMathSymbol{\varclubsuit}{\mathord}{txsymbolsC}{112}
\DeclareMathSymbol{\varclubsuitraw}{\mathord}{txsymbolsC}{112}
\newcommand{\hclubsuit}{\mathord{\mathchoice
  {\scalebox{1.4}{$\displaystyle\varclubsuitraw$}}
  {\scalebox{1.4}{$\textstyle\varclubsuitraw$}}
  {\scalebox{1.4}{$\scriptstyle\varclubsuitraw$}}
  {\scalebox{1.4}{$\scriptscriptstyle\varclubsuitraw$}}}}

\newcommand{\methodname}{EVA-0\xspace}

\definecolor{oursrow}{HTML}{E3F1EE}
\definecolor{avgcol}{HTML}{EAEFF4}
\definecolor{oursavg}{HTML}{F6DDD2}

\usetikzlibrary{backgrounds}
\definecolor{badgedark}{HTML}{2B2B2B}
\definecolor{badgelight}{HTML}{E9EDF2}

\definecolor{cnumgray}{HTML}{2B2B2B}
\DeclareRobustCommand{\cnum}[1]{\tikz[baseline={([yshift=0.08pt]c.south)}]\node[circle,draw=cnumgray,fill=cnumgray!8,line width=0.3pt,inner sep=0pt,minimum size=6.84pt,text=black,font=\fontsize{6.9pt}{6.9pt}\selectfont\rmfamily\bfseries](c){#1};}
\title{\methodname: Test-Time Model Evolution with \\ Only \textit{Two} Forward Passes per Sample}
\author{
    \vspace{-0.01in}
    Guohao Chen\textsuperscript{\rm 1}\thanks{Equal contributions. \textsuperscript{\ensuremath{\dagger}}Corresponding author. ~~Emails:~\{guohao.chen, shuaicheng.niu, jianfei.yang\}@ntu.edu.sg} ~~
    Shuaicheng Niu\textsuperscript{\rm 1,2}\footnotemark[1] ~~
    Geng Li\textsuperscript{\rm 1} ~~
    Yunbei Zhang\textsuperscript{\rm 3} ~~
    Shilin Shan\textsuperscript{\rm 1} \\
    \textbf{Chunyan Miao}\textsuperscript{\rm 1} ~~
    \textbf{Jianfei Yang}\textsuperscript{\rm 1}\textsuperscript{\ensuremath{\dagger}}\\
    \textsuperscript{\scriptsize{\rm 1}}\small{Nanyang Technological University,}
    \textsuperscript{\scriptsize{\rm 2}}\small{Jinan-NTU Green Technology Research Institute,}
    \textsuperscript{\rm 3}\small{Tulane University}
    \\
    {\normalfont\footnotesize\sffamily\hypersetup{pdfborder={0 0 0}}\textcolor[HTML]{555555}{\faGithub~\href{https://cascol-chen.github.io/EVA-0-project-page}{cascol-chen.github.io/EVA-0}}}
    \vspace{-0.14in}
}

\begin{document}

\maketitle

\begin{abstract}

Test-time model evolution offers a promising way for deployed models to improve from unlabeled test-time experience, yet most existing methods depend on backpropagation (BP), which incurs substantial memory overhead and makes them difficult to deploy on edge devices, quantized models, specialized accelerators, or black-box models. In this work, we \textbf{study test-time model evolution under a strict two-forward budget}, a setting that pushes adaptation toward highly efficient real-world deployment. We reveal three key obstacles in zeroth-order test-time optimization: susceptibility to shortcut solutions, uncontrolled weight drift, and ineffective update direction estimation. To overcome them, we propose EVA-0, a minimal zeroth-order adaptation framework that: 1) keeps the loss scale-invariant to prevent shortcut solutions; 2) devises an anchor-guided optimization strategy to alleviate weight drift; 3) uses sample-wise symmetric two-sided perturbation for update direction estimation and inference. EVA-0 requires no BP and performs both inference and adaptation within only two forward passes per sample. Results on ImageNet-C\&ViT-Base show that EVA-0 outperforms both BP-based DeYO and BP-free FOA, while achieving a 14$\times$ speed-up over FOA. 
% \href{https://github.com/Cascol-Chen/EVA-0-arxiv-}{\textit{Code is available}}.
% \textit{Code will be released.}
% shortcut exploitation of test-time objectives,
% and 3) uses symmetric sample-wise two-sided sampling to estimate more reliable update directions, while averaging the two-sided outputs for inference and thus saving one forward pass for original model inference. EVA-0 requires no backpropagation and performs both inference and adaptation within only two forward passes per sample. Extensive experiments show that EVA-0 enables effective, efficient, and stable model evolution from unlabeled test-time experience under practical deployment constraints. \textit{Source code will be released.}

\end{abstract}

\section{Introduction}

% Modern AI models are increasingly deployed in dynamic environments that continuously evolve and vary due to distribution shifts, user behavior, sensor variations, and changing real-world conditions~\cite{hendrycks2019benchmarking,koh2021wilds}. Such deployment exposes models to open-ended experience across diverse used scenarios, which provides a rich source of information unavailable during source training. However, a pretrained model that is kept frozen cannot meaningfully translate its test-time experience into improved behavior, and quickly becomes suboptimal as the environment evolves. Thus, a natural path toward more adaptive intelligence is to enable deployed models to \textit{learn from their own experience after release}~\cite{dupoux2026ai,silver2025welcome}.

% This motivates the study of test-time adaptation (TTA)~\cite{liang2023ttasurvey,wang2021tent,wang2022cotta}: enabling deployed models to update themselves online from their unlabeled test-time experience. However, most existing TTA methods remain difficult to deploy broadly as they rely on backpropagation (BP) to compute gradient~\cite{niu2024foa}, which introduces substantial memory overhead, and can be incompatible with resource-limited edge devices, quantized models, specialized accelerators, or black-box model access settings.

Next-generation AI is expected to advance not only through larger pre-training, but also by \textit{learning from its own experience} after release~\cite{silver2025welcome,dupoux2026ai}. In deployment, such experience is abundant: models constantly encounter data from changing conditions, sensors, and users~\cite{hendrycks2019benchmarking,koh2021wilds}, which are unseen in training and often observed only once. A frozen model simply discards this experience and becomes suboptimal as conditions shift. This calls for test-time model evolution, where a deployed model continually improves itself from its unlabeled test stream and remains stable over long horizons. As\linebreak such evolution must happen wherever the model runs, \textit{learning should be nearly as cheap as inference}.

% Test-Time Adaptation (TTA)~\cite{liang2023ttasurvey,wang2021tent,wang2022cotta} is the most widely studied form of this capability: adapting a model online with unlabeled test-time experience under distribution shifts. However, most existing TTA methods remain difficult to deploy broadly as they rely on backpropagation (BP) to compute gradient~\cite{niu2024foa}, which introduces substantial memory overhead, and can be incompatible with resource-limited edge devices, quantized models, specialized accelerators, or black-box model access settings.

Test-Time Adaptation (TTA)~\cite{liang2023ttasurvey,wang2021tent,wang2022cotta} is the most widely studied form of this capability: adapting a model online with unlabeled test-time experience under distribution shifts. However, most TTA methods remain difficult to deploy as they rely on backpropagation (BP) to compute gradients~\cite{niu2024foa}, which makes each update far more costly than inference, and can be incompatible with resource-limited edge devices, quantized models, specialized accelerators, or black-box model access settings.

% Our goal is not merely to improve TTA performance, but to understand and design reliable optimization mechanisms for models that must learn after deployment without gradients.

% Recent work has begun to explore BP-free TTA. FOA~\cite{niu2024foa} conducts forward-only adaptation through prompt evolution with CMA-ES~\cite{hansen2016cma}, which avoids backpropagation and model-weight updates. However, its adaptation is restricted to a low-dimensional prompt space and often requires multiple (\eg, 28) forward evaluations per sample for stable and effective learning. ZOA~\cite{deng2025zoa} extends further by applying zeroth-order (ZO) optimization to quantized networks under a two-forward budget with one-sided SPSA~\cite{spall2002multivariate}. Yet, its coarse and noisy update signal still leaves a clear performance gap, and the feature alignment loss used in both FOA and ZOA relies on source statistics, which may limit their applicability. Since most ZO methods are designed for the offline supervised setting where many forward passes are affordable, it remains unclear whether a deployed model can \textit{stably and effectively evolve from unlabeled test-time experience using only two forward passes per sample?}.
Recent works have begun to explore BP-free TTA. FOA~\cite{niu2024foa} conducts forward-only adaptation through prompt evolution with CMA-ES~\cite{hansen2016cma}. ZOA~\cite{deng2025zoa} extends further by applying zeroth-order (ZO) optimization to quantized networks under a two-forward budget with one-sided SPSA~\cite{spall2002multivariate}. However, these solutions either require many forward evaluations or produce less effective update directions under a strict two-forward budget, resulting in low efficiency or limited performance. 
Thus, it remains unclear whether a deployed model can \textit{stably and effectively evolve from unlabeled test-time experience using only two forward passes per sample?}

% two forward passes are far more informative than they appear when used through \textbf{sample-wise symmetric probing}

\begin{figure*}[t]
\centering
\vspace{-0.1in}
\includegraphics[width=0.95\linewidth]{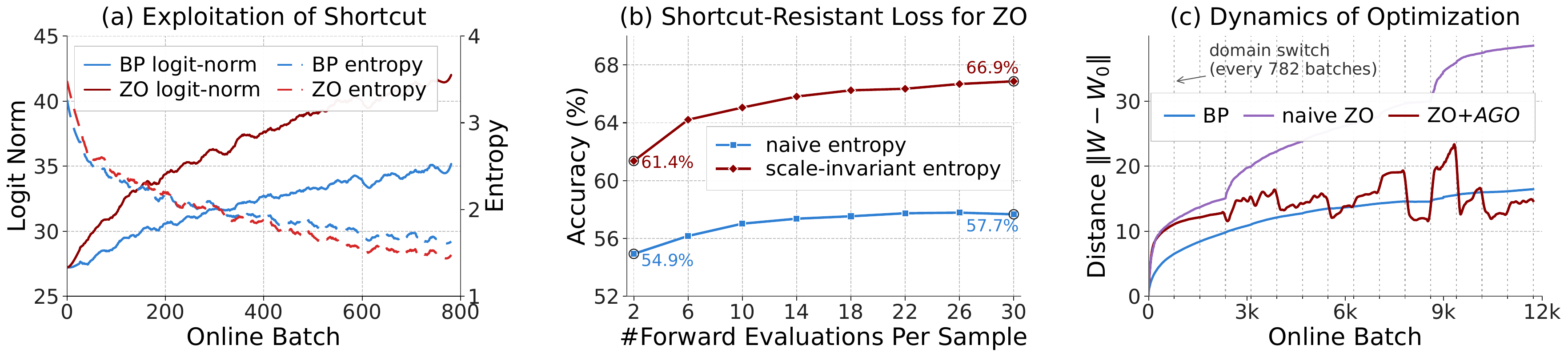}
\caption{(a) ZO optimization is more vulnerable than BP to shortcut learning: reducing entropy via logit-norm inflation. (b) Scale-invariant objective mitigates the shortcut issue and boosts performance. (c) ZO updates suffer from weight drift, and anchor-guided optimization (AGO) is able to control such drift. Results obtained on ViT-Base\&ImageNet-C, 15 corruptions for (b-c) and Gaussian for (a).}
% Optimization challenges and opportunities in ZO TTO. 
\vspace{-0.1in}
\label{fig:motivation_1_2}
\end{figure*}

To answer this question, we revisit ZO test-time optimization under the two-forward constraint through analytical and empirical studies, revealing both its key challenges and discovering new opportunities for advanced methodology design. From Figures~\ref{fig:motivation_1_2}--\ref{fig:motivation_3}, we have following key takeaways:

% \textcolor[HTML]{2F5D8C}{\ding{71}}

$\bullet$  ZO optimization is more prone to learn shortcut solutions than backpropagation (BP) under test-time loss functions, while keeping the loss scale-invariant can help mitigate this issue (c.f. Sec.~\ref{sec:objective_motivation}).

$\bullet$  Unlike BP, random ZO updates tend to cause weight drift. By properly controlling the optimization trajectory to alleviate such drift, ZO test-time learning can be better stabilized (c.f. Sec.~\ref{sec:weight_drift_motivation}).

% $\bullet$  Averaging two symmetrically perturbed predictions approximates the unperturbed prediction. Thus, one can save one forward pass for original-model inference while leveraging the more reliable update direction of two-sided SPSA using only two forward passes. In addition, sample-level perturbations can improve the diversity of sampled directions over batch-level perturbations (c.f. Sec.~\ref{sec:direction_motivation}).

% $\bullet$  Two forward passes are far more informative than they appear. First, averaging two symmetrically perturbed predictions approximates the original prediction, so one can save the extra pass for inference and use both passes for a more reliable two-sided update. Second, a separate perturbation per sample improves the diversity of update directions and their alignment with BP gradients (c.f. Sec.~\ref{sec:direction_motivation}).

$\bullet$  \textit{Two} forward passes are far more informative than they appear. Averaging two symmetrically perturbed predictions approximates the original prediction, so one can save the extra pass for inference and use both for a more reliable two-sided update. Crucially, a separate perturbation per sample improves the diversity of update directions and their alignment with BP gradients (c.f. Sec.~\ref{sec:direction_motivation}).

% To answer this question, we first revisit what makes forward-only model evolution difficult under the two-forward constraint. As in Figures~\ref{fig:motivation_1_2} \& \ref{fig:motivation_3}, our analysis highlights three key failure modes;
% \textbf{(i) inaccurate direction estimation}: one-sided SPSA exhibits a worse trade-off between inference reliability and learning efficacy than two-sided SPSA, while conventional batch-wise perturbations underuse the limited forward budget by exploring only one direction per mini-batch; 
% \textbf{(ii) exploitation of objective shortcuts}: compared with backpropagation, zeroth-order optimization is more prone to exploiting shortcut directions in test-time objectives, such as reducing entropy through inflating logit norm rather than meaningful learning; 
% \textbf{(iii) uncontrolled parameter drift}: fully random zeroth-order updates cause parameters to fluctuate and diverge over online streams, degrading generalization and stability. These findings suggest that effective forward-only model evolution is highly non-trivial.
% % These findings suggest that stable test-time model evolution requires jointly improving direction estimation, objective design, and update control.

Inspired by the above analyses, we propose \textbf{\methodname}, short for \textbf{Ev}olutionary \textbf{A}daptation with \textbf{Zero}th-Order gradients, a minimal framework for stable test-time model evolution under the strict two-forward constraint. \methodname is designed from three principles: \textbf{1) shortcut-resistant optimization:} \methodname employs \textit{scale-invariant test-time objectives}, which prevent ZO updates from reducing entropy through logit-norm inflation and encourage meaningful test-time learning; \textbf{2) controlled optimization trajectory:} \methodname introduces \textit{anchor-guided optimization}, which mitigates drift by periodically steering exploration relative to a reliable anchor, thereby enabling stable and continual model evolution; \textbf{3) accurate and informative update direction estimation:} \methodname uses symmetric \textit{two-sided SPSA} to obtain lower-variance update signals while averaging perturbed outputs for inference. It further proposes \textit{sample-wise independent perturbation} to enhance the diversity of explored update directions that significantly improves the alignment with BP update. All these mechanisms operate within the same two forward passes per sample, and \methodname requires no backpropagation.  As a result, \methodname enables effective, efficient, and stable model evolution from unlabeled test-time experience. 

\textbf{Main contributions}
1) We study test-time model evolution under a strict two-forward budget, a practical and challenging setting that pushes test-time learning/adaptation toward highly efficient real-world deployment at scale.
2) We systematically analyze the limitations of current ZO test-time optimization, revealing key challenges and design opportunities in shortcut exploitation, weight drift, and limited or biased direction estimation.
3) We propose \methodname, a two-forward test-time evolution framework that addresses these challenges with scale-invariant shortcut-resistant optimization, anchor-guided trajectory control, and sample-wise two-sided perturbation sampling and inference.
4) We validate \methodname across CNNs, ViTs, quantized models, continual streams, and black-box prompt adaptation, demonstrating effective model evolution with a practical accuracy-efficiency trade-off.

\section{Preliminary and Problem Statement}

\textbf{Test-Time Adaptation (TTA)}
Let $f_{\theta_0}: \mathcal{X}\rightarrow \mathbb{R}^C$ be a source model trained on labeled source data. During deployment, the model receives unlabeled test mini-batches $\{\mathcal{B}_t\}_{t=1}^{T}$ from a potentially shifted target distribution. Online TTA updates $f_{\theta_0}$ by optimizing an unsupervised test-time objective:
\begin{equation}\label{eq:tta-update}
    \theta_{t+1}
    =
    \theta_t - \eta \hat{g}_t,
    \qquad
    \hat{g}_t \approx \nabla_{\theta}
    \mathcal{L}(\theta_t;\mathcal{B}_t),
\end{equation}
where $\mathcal{L}$ can be entropy minimization~\cite{wang2021tent} or consistency~\cite{zhang2021memo}, \etc~Most prior methods obtain $\hat{g}_t$ via backpropagation, which might be impractical in resource-constrained or non-differentiable settings.

% which requires activation storage and gradient computation during deployment and might be impractical in resource-constrained or non-differentiable settings.

\textbf{Test-Time Forward Optimization} 
has been recently proposed to avoid BP.
% To avoid BP, recent methods perform test-time updates using only forward evaluations. 
At each step, the learner evaluates perturbed models
$
    \mathcal{Q}_t
    =
    \left\{
    f_{\theta_t+\delta_k}(\mathcal{B}_t)
    \right\}_{k=1}^{K},
$
computes the corresponding unsupervised losses, and updates $\theta_t$ only from these forward signals. 
FOA~\cite{niu2024foa} pioneers this direction by using CMA-ES~\cite{hansen2016cma} to evolve input prompts. However, CMA-ES typically requires many forward evaluations, \eg, $K=28$, and is mainly suitable for low-dimensional prompt spaces due to its high covariance estimation cost, limiting its scalability.  ZOA~\cite{deng2025zoa} adopts one-sided SPSA~\cite{spall2002multivariate} for this goal:
\begin{equation}\label{eq:one-sided-spsa}
    \widehat{g}^{\mathrm{1s}}_t
    =
    \frac{
    \ell(\theta_t+\mu z_t;\mathcal{B}_t)
    -
    \ell(\theta_t;\mathcal{B}_t)
    }{\mu}
    z_t,
    \qquad
    z_t \sim \mathcal{D},
\end{equation}
where $z_t$ is a random perturbation and $\mu$ is the perturbation scale. Although ZOA enables updates in higher-dimensional parameter spaces, its estimated update direction is still coarse and noisy, leaving a substantial room for performance improvement. Moreover, both FOA and ZOA rely on feature alignment,which requires source statistics that may be unavailable due to concerns such as privacy.

\textbf{Two-Forward \underline{Test-Time Optimization (TTO)}}
We study \textit{test-time model evolution} under a strict two-forward constraint, to ensure high run-time and memory efficiency for real-world deployment. At each online step, the model receives an unlabeled mini-batch $\mathcal{B}_t$ (the batch size could be 1) and is allowed only two forward passes per sample, without backpropagation. 
% Under this constraint, the key challenge is to obtain reliable update signals while preserving stable predictions and model evolution. 
We build our method on SPSA and address the limitations of prior one-sided forward-only adaptation.

% \section{What Hinders Two-Forward Model Evolution from Test Data?}
\section{A Closer Look at Two-Forward TTO: Challenges and Opportunities}
\label{sec:observations}

% We first examine why model evolution from unlabeled test data remains difficult for zeroth-order optimization under the strict two-forward constraint. We focus on three aspects that determine the success of forward-only evolution: how the update direction is estimated, what objective the update follows, and how the model behaves after online updates. Our analysis reveals four key observations. \blue{todo}

% \subsection{Objective Shortcuts in Zeroth-Order Test-Time Optimization}
% \subsection{Why Is Zeroth-Order TTO Difficult from an Optimization Perspective?}
\subsection{On the Learning Objective}\label{sec:objective_motivation}

\textbf{\textit{Challenge:} Zeroth-order (ZO) optimization is more prone than backpropagation (BP) to learning shortcut solutions with test-time learning objectives.}
Existing forward-only TTA methods often reuse objectives designed for BP-based adaptation in ZO updates, assuming BP-effective losses also work well for ZO. However, we show that BP and ZO differ in their sensitivity to shortcut solutions, where the test-time objective is naively reduced without improving task-relevant generalization. In particular, ZO is more susceptible to such shortcuts than BP, as formalized by
% Most objectives are developed for backpropagation-based adaptation, yet forward-only methods often reuse the same losses for zeroth-order updates. 
% This implicitly assumes that an objective useful for BP remains equally suitable when optimized only through random forward evaluations. 
% % We find that this assumption does not always hold, as BP and ZO exhibit different sensitivity to shortcuts.
% We find that this assumption does not always hold, as BP and ZO exhibit different sensitivity to shortcuts---update directions that naively decrease the test-time objective without improving task-relevant generalization.

\begin{proposition}[\textbf{\textit{Vulnerability of ZO to Shortcut Signals}}]\label{prop:zo-vulnerability}
Let $e_s$ be a unit shortcut direction and decompose the local gradient as $g=Ae_s+g_m$, where $g_m\perp e_s$. Consider the Gaussian ZO estimator $\hat{g}_{\mathrm{ZO}}=(g^\top z+\xi)z$, with $z\sim\mathcal{N}(0,I)$ and zero-mean loss noise $\xi$. For any unit direction $v\perp e_s$,
\[
    \mathbb{E}\!\left[v^\top \hat{g}_{\mathrm{ZO}}\right]=v^\top g_m,
    \qquad
    \mathrm{Var}\!\left(v^\top \hat{g}_{\mathrm{ZO}}\right)\ge A^2 .
\]
Indeed,
$
    v^\top \hat{g}_{\mathrm{ZO}}=
    A(e_s^\top z)(v^\top z)
    +
    (g_m^\top z)(v^\top z)
    +
    \xi(v^\top z).
$
The first term is induced solely by the shortcut signal, which has zero mean but variance $A^2$. Thus, the shortcut signal injects large variance into orthogonal directions related to meaningful updates, and may even flip meaningful updates. In contrast, BP directly accesses $v{^\top} g{=}v^\top g_m$ without mixing shortcut signals into orthogonal directions.
\end{proposition}

% Proposition~\ref{prop:zo-vulnerability} suggests that ZO is more affected by objective shortcuts than BP. 

\textbf{\textit{Opportunity:} Keeping the entropy objective scale-invariant could help mitigate shortcut learning.}
We further empirically study the above proposition using the widely used entropy loss~\cite{wang2021tent,niu2023sar,lee2024deyo}. Entropy can be reduced either by improving the decision boundary or by simply inflating the logit norm, the latter forming a shortcut direction without meaningful learning. As in Figure~\ref{fig:motivation_1_2} (a), when BP and ZO achieve similar entropy reduction, ZO induces much larger logit norms, indicating its stronger susceptibility to such shortcuts (details in App.~\ref{suppl:sec:more-discussion}). Moreover, as in Figure~\ref{fig:motivation_1_2} (b), making entropy logit-norm-invariant (\ie, Eqn.~\eqref{eq:shortcut_resistant_entropy} without de-centering) significantly improves ZO performance. Thus, two-forward model evolution requires objectives that are not only effective but also shortcut-resistant.

\subsection{On the Weight Dynamics}\label{sec:weight_drift_motivation}

% \paragraph{Observation 4: random zeroth-order updates accumulate into weight drift.}
% Unlike BP, whose update direction is determined by the mini-batch gradient on $\mathcal{B}_t$, ZO first samples a random perturbation direction and only uses a scalar loss difference to decide how to move along it. As a result, each update contains useful descent components as well as high-dimensional random components that are unrelated to stable model evolution, making the parameter trajectory noisy and oscillatory.

% Once a noisy update overfits or pushes the model toward an erroneous region, later SPSA updates are unlikely to reverse that mistake: in high-dimensional spaces, a random perturbation is nearly orthogonal to any fixed recovery direction toward a stable solution. This creates a drift mechanism in ZO: \underline{\textit{short-term loss reduction may coexist with long-term parameter deviation}}. As in Figure~\ref{fig:obs_drift}(a), naive ZO updates lead to growing parameter norms and eventually degrade performance, whereas BP-based updates remain more stable under the same objective. This suggests that, in ZO, the model should remain plastic enough to adapt to test-time evidence, while its trajectory should be controlled.

\textbf{\textit{Challenge and Opportunity:} Random ZO updates can accumulate into weight drift, whereas BP updates are less prone to this issue.}
Unlike BP, ZO updates rely on random perturbation directions scaled by loss differences. Thus, even when a step reduces the current test-time loss, it may include components unrelated to stable model improvement, especially under noisy unsupervised objectives. These components can accumulate over online streams and move the model away from the pretrained stable region. Since later perturbations are independently sampled, they are unlikely to align with the recovery direction in high-dimensional space. As in Figure~\ref{fig:motivation_1_2} (a), naive ZO lead to growing weight norms and degraded performance (Table~\ref{tab:eva0-ablation}), whereas BP-based updates remain more stable under the same objective. This motivates trajectory control in addition to objective minimization.

\subsection{On the Gradient Direction Estimation Strategy}\label{sec:direction_motivation}

\begin{figure*}[t]
\centering
% \vspace{-0.1in}
\includegraphics[width=0.95\linewidth]{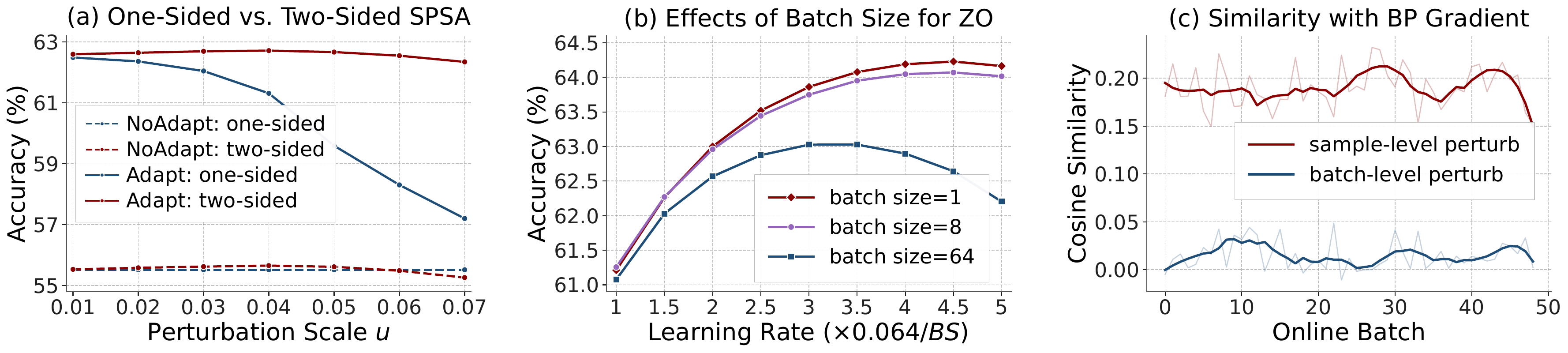}
\caption{ 
% (a) Accuracy comparison of different direction estimation and inference strategies. \textit{One-sided} SPSA uses the original model output for inference, while \textit{two-sided} SPSA uses the averaged outputs from two symmetrically perturbed models, which approximates clean-model inference.
% (b) Effects of different batch sizes in ZO test-time optimization. Smaller batches provide more independent perturbation samples, yielding more diverse update directions and better performance.
% (c) Similarity between ZO and BP gradients under different perturbation strategies. Sample-wise perturbations produce ZO update directions that are better aligned with BP gradients.
(a) Accuracy of different direction estimation and inference strategies. \textit{One-sided} SPSA uses the original output for inference, while \textit{two-sided} SPSA averages two perturbed outputs to approximate clean inference.
(b) Smaller batches provide more independent perturbations, yielding more diverse update directions and better performance.
% Effects of batch size in ZO test-time optimization.
(c) Sample-wise perturbations yield ZO directions better aligned with BP gradients.
% Similarity between ZO and BP gradients. 
Results obtained on ViT-Base \& ImageNet-C (15 corruptions for (a-b) and Gaussian for (c)) using CE loss.
% Results obtained with ViT-Base on ImageNet-C using CE loss, (c) uses $BS$ of 1024. (a-b all corruptions, c Gauss, but this is not important for this figure)
}
\label{fig:motivation_3}
\end{figure*}

\textbf{\textit{Challenge:} One-sided SPSA is sensitive to perturbation noise, while two-sided SPSA requires more forward passes.}
Existing forward-only TTA methods typically need to allocate one forward pass to the original model for inference. Thus, under a strict two-forward budget, one has only a single perturbed evaluation (one forward pass) for estimating the update direction, as one-sided SPSA in Eqn.~(\ref{eq:one-sided-spsa}). However, as in Figure~\ref{fig:motivation_3} (a), one-sided SPSA preserves stable inference but makes adaptation effectiveness limited and sensitive to perturbation noise. In contrast, two-sided SPSA uses the two evaluations as a symmetric local probe around the current model:
\begin{equation}\label{eq:two-sided-spsa}
\widehat{g}^{\mathrm{2s}}_t
=
\frac{
\ell(\theta_t+\mu z_t;\mathcal{B}_t)
-
\ell(\theta_t-\mu z_t;\mathcal{B}_t)
}{2\mu} z_t ,
\qquad z_t \sim \mathcal{D}.
\end{equation}
This usually yields a more reliable update direction, but it appears to require an additional clean forward pass for inference, \ie, at least three forward passes, violating the two-forward constraint.

\textbf{\textit{Opportunity:} Averaging two-sided probing outputs approximates clean-model inference without an additional forward pass.}
To ensure two-forward efficiency, unlike standard uses of two-sided SPSA merely as a variance-reduction estimator, we further exploit its symmetric outputs for inference. Under a local smoothness assumption, a Taylor expansion around $\theta_t$ gives
\begin{equation}
    f_{\theta_t \pm \mu z_t}(x)
    =
    f_{\theta_t}(x)
    \pm
    \mu J_{\theta_t}(x) z_t
    +
    O(\mu^2).
\end{equation}
Thus, averaging $f_{\theta_t+\mu z_t}(x)$ and $f_{\theta_t-\mu z_t}(x)$ cancels the first-order term and yields a second-order approximation to clean-model inference. This reveals a useful opportunity for two-forward test-time evolution: the same two symmetric evaluations can provide both a lower-variance update direction and a reliable inference output, without requiring an additional clean forward pass. Empirically, Figure~\ref{fig:motivation_3} (a) confirms that averaged symmetric predictions remain stable across perturbation scales, while the resulting update signals lead to consistently stronger adaptation.

% \paragraph{Observation 2: batch-level perturbation underuses the limited forward budget.}

% \paragraph{Opportunity: Batch-level perturbations waste sample-level directional diversity.}
\textbf{\textit{Opportunity:} Sample-level perturbations improve directional diversity than shared batch-level perturbations.}
Applying a \textit{shared perturbation} across a mini-batch is a common practice in ZO optimization, as in Eqns.~(\ref{eq:one-sided-spsa}-\ref{eq:two-sided-spsa}). Recent ZO studies further suggest using \textit{large batches} to reduce loss variance and stabilize offline training~\cite{gautam2024variance}. However, under two-forward constraint, a mini-batch with a shared perturbation explores only one direction in parameter space, regardless of the batch size.

In the online learning setting, where each sample is used only \textit{once}, we observe a phenomenon that contrasts with prior studies: smaller batches often lead to better adaptation. As in Figure~\ref{fig:motivation_3} (b), reducing the batch size consistently improves performance across learning rates. This suggests that, in streaming ZO TTO, \underline{\textit{directional exploration can be more valuable than batch-level}} \underline{\textit{loss averaging}}.
To directly measure this effect, we use the backpropagation gradient on the same mini-batch as an oracle reference and compare its similarity to the zeroth-order update direction. As in Figure~\ref{fig:motivation_3} (c), \textit{independent perturbations} across samples yield substantially higher gradient similarity than a shared batch-level perturbation, \ie, increasing similarity with BP gradient from \textit{0.015} to \textit{0.190}. This provides direct evidence that test samples can contribute useful directional information beyond scalar loss averaging, motivating sample-wise directional exploration under the two-forward constraint.

\begin{figure*}[t]
\centering
% \vspace{-0.1in}
\includegraphics[width=1\linewidth]{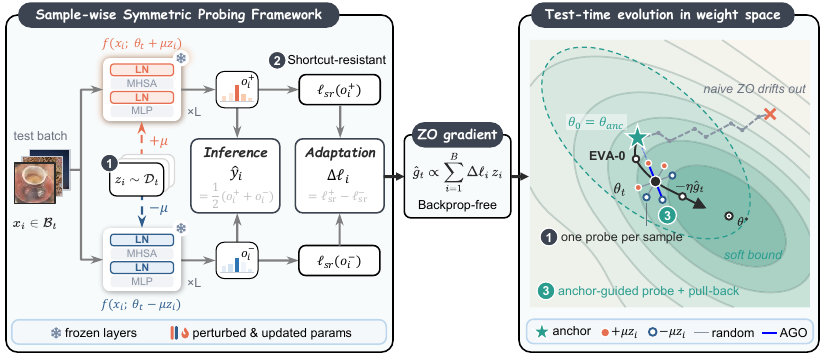}
\caption{ 
% An illustration of our proposed \methodname. For each test sample $x_i$, we draw an independent perturbation $z_i\sim\mathcal{D}_t$ and run two symmetric forward passes $f(x_i;\theta_t\pm\mu z_i)$ (\ding{182}), whose averaged output gives the prediction and whose shortcut-resistant loss difference $\Delta\ell_i$ (\ding{183}) yields a backpropagation-free gradient $\hat{g}_t\propto\sum_{i}\Delta\ell_i z_i$. Guided by the anchor $\theta_{\mathrm{anc}}$ (\ding{184}), EVA-0 evolves toward the target optimum $\theta^{*}$ within a soft bound, while naive ZO updates can drift away.
An illustration of our proposed \methodname. For each test sample $x_i$, we draw an independent perturbation $z_i\sim\mathcal{D}_t$ and run two symmetric forward passes $f(x_i;\theta_t\pm\mu z_i)$ (\cnum{1}), whose averaged output gives the prediction and whose shortcut-resistant loss difference $\Delta\ell_i$ (\cnum{2}) yields a backpropagation-free gradient $\hat{g}_t$. We then update $\theta_t$ under anchor-guided optimization (\cnum{3}), so that \methodname evolves toward the optimum within a soft bound, while naive ZO updates drift away.
}
\label{fig:eva0-method}
\end{figure*}

\section{\methodname: Stable and Effective Two-Forward Model Evolution at Test Time}
\label{sec:method}

In this section, we propose \textbf{\methodname}, short for \textbf{Ev}olutionary \textbf{A}daptation with \textbf{Zero}th-order gradients. \methodname aims to achieve stable and effective two-forward test-time model evolution by addressing the challenges revisited in Sec.~\ref{sec:observations}. Motivated by the opportunities discovered in Sec.~\ref{sec:observations}, \methodname instantiates them as three design principles: 1) sample-wise two-sided perturbation for direction estimation and inference (c.f. Sec.~\ref{sec:method-direction}); 2) shortcut-resistant objective (c.f. Sec.~\ref{sec:method-objective}); 3) bounded yet plastic online optimization (c.f. Sec.~\ref{sec:method-anchor}),  and then integrates them into a simple, unified 2-forward ZO update rule:
\begin{equation}
\label{eq:eva0-gradient}
    \hat{g}_t^{\mathrm{EVA}}
    =
    \frac{1}{B}
    \sum_{i=1}^{B}
    \frac{
    \ell_{\mathrm{sr}}\!\left(\theta_t+\mu z_{t,i}; x_i\right)
    -
    \ell_{\mathrm{sr}}\!\left(\theta_t-\mu z_{t,i}; x_i\right)
    }{2\mu}
    z_{t,i},
    \qquad
    z_{t,i}\sim \mathcal{D}_t ,
\end{equation}

Here, $\ell_{\mathrm{sr}}$ denotes a \textit{shortcut-resistant} test-time objective to encourage meaningful learning, and $\mathcal{D}_t$ is an adaptive rather than fixed perturbation distribution that co-evolves with $\theta_t$ to control the model trajectory. The overall \textbf{pseudo-code of \methodname is summarized in Algorithm~\ref{alg:eva0} in Appendix~\ref{appx:sec:algorithm}. }

\subsection{Sample-wise Symmetric Update Direction Estimation and Inference \cnum{1}}
\label{sec:method-direction}

% As shown in Sec.~\ref{sec:direction_motivation}, standard SPSA designs fall short under the two-forward constraint. One-sided SPSA preserves clean inference but sacrifices adaptation quality, while the prior two-sided SPSA requires at least three forward passes. \methodname addresses this hurdle through two-sided sampling with an averaged output of two-sided perturbed predictions. Meanwhile, considering sample-level perturbations provides more diverse update directions diversity than shared batch-level perturbations as we illustrated in Sec.~\ref{sec:direction_motivation}, We therefore sample perturbations independently for each sample. Overall, we name this strategy as \textit{sample-wise symmetric direction estimation} (SSD). For each sample $x_i$, we sample an independent perturbation $z_{t,i}\sim\mathcal{D}_t$ and evaluate two symmetric models,

Following Sec.~\ref{sec:direction_motivation}, \methodname uses \textit{sample-wise symmetric direction estimation} (SSD). For each sample $x_i$, we draw an independent perturbation $z_{t,i}\sim\mathcal{D}_t$ and evaluate two symmetric models:
\[
    f_{\theta_t+\mu z_{t,i}}(x_i)
    \quad\text{and}\quad
    f_{\theta_t-\mu z_{t,i}}(x_i).
\]
The two forward passes are used for both inference and adaptation. \textbf{For adaptation}, \methodname uses the two-sided loss difference, as defined in Eqn.\eqref{eq:eva0-gradient}. \textbf{For inference}, \methodname averages the two perturbed outputs, $f_{\theta_t\pm\mu z_{t,i}}(x_i)$, which provides a second-order approximation to the clean output under local smoothness, as shown in Figure~\ref{fig:motivation_3} (a).

This symmetric strategy satisfies the two-forward constraint while retaining the more reliable direction estimation enabled by two-sided perturbation. The sample-wise perturbations further turn each mini-batch into \textit{multiple directional probes}, rather than a single loss-averaging unit. As a result, EVA-0 obtains update directions that are both more reliable and more informative from streaming test samples. We provide a memory- and computation-efficient implementation of SSD in Appendix~\ref{suppl:sec:efficient-implementation}.

\subsection{Shortcut-Resistant and Sample-Wise Test-Time Objectives \cnum{2}}
\label{sec:method-objective}

As discussed in Sec.~\ref{sec:objective_motivation}, ZO optimization is more prone than BP to learning shortcut solutions, and maintaining a scale-invariant objective can help mitigate this issue. Therefore, in this section, we introduce two scale-invariant objectives for two-forward TTO. The first works in the fully source-free TTA setting, while the latter can further improve performance when source statistics are available.

% Observation 3 shows that ZO optimization based on scalar feedback is more vulnerable to objective shortcuts. Meanwhile, sample-wise perturbation requires the objective to remain valid even when each perturbation is evaluated on a single sample. Therefore, EVA-0 designs $\ell_{\mathrm{sr}}$ to be both \textit{shortcut-resistant} and \textit{sample-wise compatible}. Following FOA, we use entropy minimization and feature-statistic alignment as the basic test-time signals, and adapt both to sample-wise two-forward evolution.

\textbf{Shortcut-resistant entropy minimization}
Vanilla entropy minimization is effective for TTA, but it admits shortcut solutions, which become more severe from BP to ZO optimization: the loss can be reduced by inflating logit norms or collapsing predictions to a few high-confidence classes. EVA-0 mitigates these shortcuts via scale normalization and output decentering. Given symmetric logits $o^\pm_{t,i}=f_{\theta_t\pm\mu z_{t,i}}(x_i)$, let $\bar{o}_{t,i}=(o^+_{t,i}+o^-_{t,i})/2$ and $\bar r_{t,i}=\|\bar{o}_{t,i}\|_2$. For $o\in\{o^+_{t,i},o^-_{t,i}\}$, we define
\begin{equation}\label{eq:shortcut_resistant_entropy}
    s(o)
    =
    \frac{\bar r_{t,i}}{\|o\|_2+\epsilon}o
    -
    c_t,
    \qquad
    p(o)=\mathrm{softmax}(s(o)),
    \qquad
    E^{\mathrm{SR}}(o)
    =
    -\sum_{k} p_k(o)\log p_k(o),
\end{equation}
where $c_t$ is an online output center maintained via EMA over $\bar{o}_{t,i}$, inspired by DINO~\cite{caron2021emerging}. The normalization term removes the logit-scale degree of freedom by evaluating both perturbed outputs at the same norm $\bar r_{t,i}$, preventing entropy reduction through norm inflation. The decentering term improves output diversity by suppressing persistent output collapse. 
% Notably, the decentering term preserves pseudo-labels of high-confident predictions, showing effectiveness even under an imbalanced target distribution (Table~\ref{xxx}). 
Together, they allow ZO to extract more meaningful updates from entropy.

\textbf{Sample-wise feature-statistic alignment}
When source feature statistics are available, feature alignment provides an additional reference for meaningful adaptation, as in FOA~\cite{niu2024foa} and ZOA~\cite{deng2025zoa}. The alignment loss is naturally more scale-invariant. However, under sample-wise perturbations (as described in Sec.~\ref{sec:method-direction}) and single-sample adaptation scenarios, it becomes ill-posed because target statistics are estimated from a single sample, yielding noisy means and ill-defined variances.

% \methodname makes this alignment well-defined by estimating how each sample updates the online target distribution. Instead of directly estimating variance, \methodname tracks the first and second moments separately over streaming test samples, and reconstructs the variance via $\mathrm{Var}[h]=\mathbb{E}[h^2]-\mathbb{E}[h]^2$. Formally, let $h_\theta(x)$ be the feature used for alignment, and let $(m_s,\sigma_s)$ be the source mean and standard deviation. EVA-0 maintains online target moments $(m_t,q_t)$, where $m_t$ tracks $\mathbb{E}[h]$ and $q_t$ tracks $\mathbb{E}[h^2]$. For a perturbed feature $h=h_{\theta_t\pm\mu z_{t,i}}(x_i)$, we define the updated moments as

\methodname makes this alignment well-defined by estimating how each sample updates the online target distribution. As one sample has no variance, \methodname instead tracks the first and second moments over the test stream and recovers the variance via $\mathrm{Var}[h]=\mathbb{E}[h^2]-\mathbb{E}[h]^2$.
Formally, let $h_\theta(x)$ be the alignment feature with source mean and standard deviation $(m_s,\sigma_s)$, and $(m_t,q_t)$ be the running estimates of $\mathbb{E}[h]$ and $\mathbb{E}[h^2]$. For a perturbed feature $h=h_{\theta_t\pm\mu z_{t,i}}(x_i)$, the updated moments are
\begin{equation}\label{eq:smooth-for-align}
    \hat{q}_t(h)=(1-\rho)q_t+\rho h^2,
    \qquad
    \hat{m}_t(h)=(1-\rho)m_t+\rho h,
    \qquad
    \hat{\sigma}_t(h)=\sqrt{\hat{q}_t(h)-\hat{m}_t(h)^2}.
\end{equation}
The sample-wise alignment loss is then computed by
\begin{equation}\label{eq:overall-objective}
    \ell_{\mathrm{SWA}}(h)
    =
    \|\hat{m}_t(h)-m_s\|_2^2
    +
    \|\hat{\sigma}_t(h)-\sigma_s\|_2^2 .
\end{equation}
This EMA formulation makes feature alignment compatible with sample-wise two-forward evolution: each sample provides a valid alignment signal, while moving moments stabilize the online estimates. After each update, $(m_t,q_t)$ are updated using the averaged symmetric features, allowing the statistics to evolve online without extra forward passes. Then, the \textbf{Overall Objective} of \methodname is given by
\begin{equation}
    \ell_{\mathrm{sr}}(\theta;x)
    =
    E^{\mathrm{SR}}(o)
    +
    \lambda\ell_{\mathrm{SWA}}(h).
\end{equation}
When source feature statistics are unavailable, we set $\lambda=0$, yielding fully test-time adaptation.

\subsection{Anchor-Guided Bounded Test-Time Evolution \cnum{3}}
\label{sec:method-anchor}

% As discussed in Sec.~\ref{sec:weight_drift_motivation}, random ZO updates can accumulate into hard-to-recover weight drift due to coarse direction estimates. Thus, test-time evolution needs not only exploration, but also a restoring mechanism to keep the trajectory controlled and the model stable.

% We seek to control the optimization trajectory prevent from weight drift, thereby improve stability.

% Observation 4 shows that fully random ZO updates can accumulate into weight drift: once a noisy update overfits the current mini-batch or moves the model toward an erroneous region, later random perturbations are unlikely to reliably correct this mistake. Therefore, test-time evolution requires not only exploration, but also a restoring mechanism that keeps the model within a stable neighborhood.

% which is hard to recover from since ZO directions are only coarsely estimated rather than precisely computed as BP gradients

% Based on Sec.~\ref{sec:weight_drift_motivation}, random ZO updates can accumulate into weight drift, which is hard to recover from later ZO updates since ZO directions are randomly sampled. Thus, test-time evolution needs not only exploration, but also a restoring mechanism to keep the evolution trajectory controlled and the model stable.
% \methodname achieves this with an adaptive perturbation distribution $\mathcal{D}_t$. 

Following Sec.~\ref{sec:weight_drift_motivation}, \methodname achieves optimization trajectory control with an adaptive perturbation distribution $\mathcal{D}_t$. 
Let $\theta^{\mathrm{anc}}$ be a reliable teacher or anchor, and $d_t=\theta^{\mathrm{anc}}-\theta_t$ denote the anchor direction. For each mini-batch $\mathcal{B}_t$ with $B$ samples, \methodname assigns $B-k$ samples to random exploration ($i\notin \mathcal{I}_t$) and remaining $k$ samples $i\in \mathcal{I}_t\subset\mathcal{B}_t$ to anchor-guided perturbations. That is,
% $z_{t,i}\sim \mathcal{D}_t$ is given by
\begin{equation}
\label{eq:anchor-distribution}
    z_{t,i}\sim\mathcal{D}_t
    =
    \begin{cases}
    \bar d_t \odot r_{t,i}, & i\in \mathcal{I}_t, \quad r_{t,i}\sim \mathcal{U}(0,2), \\
    u_{t,i}, & i\notin \mathcal{I}_t, \quad u_{t,i}\sim \mathcal{N}(0,I),
    \end{cases}
    \quad
    \text{where~}
    \bar d_t
    =
    \frac{\mathbb{E}_{u\sim\mathcal{N}(0,I)}[\|u\|_2]}
    {\|d_t\|_2+\delta}
    d_t.
\end{equation}
Here, $\odot$ is element-wise multiplication. 
Under the symmetric two-forward estimator, the two sides $\theta_t+\mu z_{t,i}$ and $\theta_t-\mu z_{t,i}$ correspond to moving relatively closer to or farther from the anchor. The shortcut-resistant loss compares these two directions and determines which one better fits the current test sample. As a result, anchor-guided perturbation provides a recovery-aware search direction that can pull the model back toward a stable region when needed, while still allowing adaptation away from the anchor when supported by test-time evidence.

The above anchor-guided perturbation controls the update direction in a loss-driven manner. Here, we further apply a weak relaxation that acts as a smoother to suppress the parameter drift:
\begin{equation}\label{eq:weak-restore}
    \theta_{t+1}
    =
    (1-\gamma)\theta_{t+1}^{\prime}
    +
    \gamma \theta^{\mathrm{anc}},
\end{equation}
where $\theta_{t+1}^{\prime}$ is an intermediate model after the ZO update.
% This relaxation acts as a slow stabilizer over the online stream.
With a small $\gamma$, the model remains plastic enough to learn from test-time evidence, while its trajectory is denoised and softly bounded that prevents drift. 
In \methodname, $\theta^{\mathrm{anc}}$ can be realized as the pretrained weights or a slowly moving EMA of the online parameters (see Table~\ref{tab:anchor-design}).

% The anchor-guided perturbation sampling
% Then, $\mathcal{D}_t$ is a mixture of random exploration and anchor-guided exploration in Eqn.~\eqref{eq:anchor-distribution}, and its anchor component evolves with the direction to  $\theta^{\mathrm{anc}}$. 

% Under the symmetric two-forward estimator, the two sides $\theta_t+\mu z_{t,i}$ and $\theta_t-\mu z_{t,i}$ compare moving relatively closer to and farther from the anchor. The shortcut-resistant loss then determines which side is preferable for the current test sample, and introduces a loss-driven update trajectory control. Thus, anchor-guided perturbation provides a recovery-aware search direction, while still allowing the model to adapt when test-time evidence supports moving away from the anchor solution.

% In addition, we also apply a weak relaxation toward the anchor after each update to further stabilize,
% Moreover, we add a weak persistent force toward the anchor to suppress the accumulation of small drift in random test-time ZO updates, while allowing consistent update signals to stand out, define as:

% While anchor-guided perturbation acts at the update-direction generation level, we further apply a weak relaxation toward the anchor at the parameter level after each update:

\begin{table}[t]
    \vspace{-0.15in}
    \caption{Comparisons with SOTA methods on ImageNet-C (severity level 5) regarding \textbf{Accuracy~(\%)} under \texttt{single-domain TTA}, \textbf{BP} is short for backward propagation, \textbf{FTTA} is short for source-free fully TTA, and \textbf{\#FP} is the number of forward passes a sample. $\dagger$ indicates a variant that uses only prediction entropy as the test-time objective, \ie, $\lambda{=}0$.
    The \textbf{bold} number indicates the best result.
    }
    \vspace{-0.1in}
    \setlength{\tabcolsep}{3.2pt}
    \label{tab:single-domain-tto}
\newcommand{\tabincell}[2]{\begin{tabular}{@{}#1@{}}#2\end{tabular}}
 \begin{center}
 \begin{threeparttable}
 \LARGE
    \resizebox{1.0\linewidth}{!}{
\begin{tabular}{clcccccccccccccccccc>{\columncolor{blue!8}}c}
    \multicolumn{5}{c}{} 
    & \multicolumn{3}{c}{Noise} 
    & \multicolumn{4}{c}{Blur} 
    & \multicolumn{4}{c}{Weather} 
    & \multicolumn{4}{c}{Digital} 
    & \multicolumn{1}{c}{} \\

    \multicolumn{2}{l}{Model+Method} & BP & FTTA  & \#FP 
    & Gaus. & Shot & Imp. 
    & Def. & Glass & Mot. & Zoom 
    & Snow & Frost & Fog & Brit. 
    & Contr. & Elas. & Pix. & JPEG & Avg. \\

    \midrule

    \multirow{11}{*}{\rotatebox[origin=c]{90}{ViT-Base}}
    & NoAdapt & \ding{55} & \ding{52} & 1 & 56.8 & 56.8 & 57.5 & 46.9 & 35.6 & 53.1 & 44.8 & 62.2 & 62.5 & 65.7 & 77.7 & 32.6 & 46.0 & 67.0 & 67.6 & 55.5 \\

    & TENT & \ding{52} & \ding{52} & 1 & 60.3 & 61.6 & 61.8 & 59.2 & 56.5 & 63.5 & 59.2 & 54.3 & 64.5 & 2.3 & 79.1 & 67.4 & 61.5 & 72.5 & 70.6 & 59.6 \\

    & CoTTA & \ding{52} & \ding{52} & 3or35 & 63.6 & 63.8 & 64.1 & 55.5 & 51.1 & 63.6 & 55.5 & 70.0 & 69.4 & 71.5 & 78.5 & 9.7 & 64.5 & 73.4 & 71.2 & 61.7 \\

    & SAR & \ding{52} & \ding{52} & [1, 2] & 59.2 & 60.5 & 60.7 & 57.5 & 55.6 & 61.8 & 57.6 & 65.9 & 63.5 & 69.1 & 78.7 & 45.7 & 62.4 & 71.9 & 70.3 & 62.7 \\

    & DeYO & \ding{52} & \ding{52} & [1, 2] & 59.8 & 61.0 & 60.2 & 58.4 & 58.8 & 64.4 & 62.3 & 68.8 & 66.4 & 70.1 & 78.8 & 62.2 & 68.4 & 74.1 & 72.2 & 65.7 \\

    & ZOA & \ding{55} & \ding{55} & 2 & 58.2 & 59.7 & 60.3 & 52.6 & 47.2 & 58.7 & 53.9 & 66.1 & 62.9 & 67.6 & 78.7 & 56.9 & 56.8 & 70.2 & 70.1 & 61.3 \\

    & FOZO & \ding{55} & \ding{55} & 2 & 58.7 & 61.1 & 62.5 & 51.1 & 44.7 & 57.8 & 53.4 & 66.3 & 66.3 & 67.8 & 79.1 & 59.8 & 55.4 & 70.4 & 71.8 & 61.7  \\

    & FOA & \ding{55} & \ding{55} & 28 & 61.5 & 63.2 & 63.3 & 59.3 & 56.7 & 61.4 & 57.7 & 69.4 & 69.6 & 73.4 & 81.1 & 67.7 & 62.7 & 73.9 & 73.0 & 66.3 \\
    & FOA$^\dagger$ & \ding{55} & \ding{52} & 28 & 58.2 & 59.6 & 59.1 & 55.1 & 36.0 & 54.0 & 47.3 & 47.8 & 59.1 & 72.5 & 77.2 & 30.0 & 48.7 & 66.6 & 67.9 & 55.9 \\

     \rowcolor{black!4}\cellcolor{white} & \methodname (ours) & \ding{55} & \ding{55} & 2 & 61.8 & 63.0 & 63.8 & 58.1 & 56.4 & 62.3 & 60.5 & 70.0 & 68.5 & 74.2 & 80.7 & 68.5 & 66.4 & 73.8 & 73.2 & \textbf{66.7} \\

      \rowcolor{black!4}\cellcolor{white} & EVA-0$^\dagger$ (ours) & \ding{55} & \ding{52} & 2 & 60.1 & 61.3 & 62.0 & 56.8 & 54.7 & 60.8 & 57.4 & 67.7 & 66.2 & 71.7 & 79.4 & 66.7 & 62.1 & 72.0 & 70.3 & 64.6 \\

    \midrule

    \multirow{11}{*}{\rotatebox[origin=c]{90}{R50-GN}}
    & NoAdapt & \ding{55} & \ding{52} & 1 & 22.1 & 23.0 & 22.0 & 19.8 & 11.4 & 21.5 & 25.0 & 40.3 & 47.0 & 34.0 & 68.8 & 36.2 & 18.5 & 29.2 & 52.6 & 31.4 \\

    & TENT & \ding{52} & \ding{52} & 1 & 24.2 & 28.0 & 25.0 & 15.2 & 7.7 & 22.0 & 22.6 & 26.9 & 33.9 & 3.5 & 69.9 & 42.1 & 11.0 & 48.6 & 54.4 & 29.0 \\

    & CoTTA & \ding{52} & \ding{52} & 3or35 & 32.3 & 34.2 & 34.6 & 12.4 & 15.9 & 13.9 & 26.4 & 39.8 & 39.6 & 7.2 & 66.7 & 23.9 & 18.1 & 50.2 & 56.8 & 31.5 \\

    & SAR & \ding{52} & \ding{52} & [1, 2] & 32.3 & 34.6 & 33.5 & 18.9 & 19.7 & 30.4 & 30.8 & 42.6 & 43.3 & 6.2 & 70.4 & 43.8 & 15.9 & 49.0 & 55.2 & 35.1 \\

    & DeYO & \ding{52} & \ding{52} & [1, 2] & 40.7 & 42.7 & 42.3 & 21.6 & 23.6 & 38.1 & 36.4 & 51.1 & 49.4 & 54.9 & 72.9 & 50.1 & 42.1 & 56.3 & 57.7 & 45.3 \\

    % & T3A & \ding{55} & \ding{52} & 1 & 56.4 & 56.9 & 57.3 & 47.9 & 37.8 & 54.3 & 46.9 & 63.6 & 60.8 & 68.5 & 78.1 & 38.3 & 50.0 & 67.6 & 69.1 & 56.9 \\

    & ZOA & \ding{55} & \ding{55} & 2 & 29.0 & 29.2 & 27.7 & 23.6 & 15.2 & 27.3 & 30.2 & 44.5 & 46.5 & 49.9 & 69.9 & 42.4 & 26.3 & 41.6 & 55.8 & 37.3 \\

    & FOZO & \ding{55} & \ding{55} & 2 & 32.1 & 32.5 & 33.7 & 13.1 & 8.6 & 18.0 & 21.3 & 38.8 & 44.3 & 31.4 & 68.0 & 35.2 & 22.8 & 51.4 & 55.3 & 33.8  \\

    & FOA & \ding{55} & \ding{55} & 28 & 27.3 & 27.6 & 23.1 & 21.1 & 11.5 & 23.7 & 26.9 & 38.6 & 48.5 & 42.0 & 70.1 & 41.6 & 20.7 & 51.4 & 55.3 & 35.3 \\
    & FOA$^\dagger$ & \ding{55} & \ding{52} & 28 & 26.9 & 27.0 & 20.7 & 21.0 & 11.2 & 23.3 & 27.0 & 39.3 & 48.1 & 42.4 & 70.0 & 41.7 & 21.4 & 51.3 & 55.0 & 35.1 \\

    % \cmidrule(l){2-21}
    
     \rowcolor{black!4}\cellcolor{white} & \methodname (ours) & \ding{55} & \ding{55} & 2 & 37.0 & 38.2 & 37.9 & 29.1 & 24.9 & 37.2 & 41.5 & 51.0 & 51.0 & 59.5 & 72.6 & 47.9 & 41.7 & 54.2 & 57.3 & \textbf{45.4} \\

      \rowcolor{black!4}\cellcolor{white} & EVA-0$^\dagger$ (ours) & \ding{55} & \ding{52} & 2 & 32.3 & 35.2 & 34.2 & 23.8 & 15.5 & 32.6 & 34.9 & 47.4 & 46.9 & 52.6 & 71.4 & 44.8 & 31.4 & 51.8 & 56.2 & 40.7 \\      
\end{tabular}
	}
	 \end{threeparttable}
\vspace{-0.1in}	 
  \end{center}
\end{table}

\section{Comparisons with State-of-the-arts}\label{sec:main_exp}

% We demonstrate that \methodname achieves a superior tradeoff between optimization effectiveness, long-term stability, and forward evaluations.
We organize the experiments to demonstrate the efficacy of \methodname across \textit{CNNs}, \textit{ViTs}, \textit{continual streams}, \textit{quantized models},  \textit{black-box models}, and provide further insights into the designs of \methodname.

\textbf{Dataset and methods} We conduct experiments on ImageNet-C~\citep{hendrycks2019benchmarking}, a large-scaled benchmark to assess out-of-distribution generalization. It contains 15 types of 4 main categories (noise, blur, weather, digital) corrupted images and each type has 5 severity levels. We benchmark \methodname against the following TTO methods: (i) BP-based methods: TENT~\cite{wang2021tent}, CoTTA~\citep{wang2022cotta}, SAR~\citep{niu2023sar}, DeYO~\citep{lee2024deyo}, which use EMA-teacher regularization, sharpness-aware optimization, and active learning for BP-based TTO;
(ii) Forward-Only methods: FOA~\cite{niu2024foa}, ZOA~\citep{deng2025zoa}, FOZO~\citep{fozo2026}, that are based on ZO.

\textbf{Models and implementation details} We use both transformer and CNN models, ViT-Base~\cite{dosovitskiy2021an} and ResNet50-GN~\cite{he2016deep}, for our experiments.
The models are trained on ImageNet-1K and the model weights are obtained from the \texttt{timm} repository~\cite{rw2019timm}. We adopt PTQ4ViT~\cite{yuan2022ptq4vit} for 8-bit and 6-bit ViT quantization. We update the affine parameters in norm layers per ZOA~\cite{deng2025zoa}, using SGD with a learning rate of 0.002. We use the pretrained weights as $\theta^{\mathrm{anc}}$ with $\gamma=0.001$ in Eqn.~(\ref{eq:weak-restore}).
Discussions on hyperparameters can be found in Sec.~\ref{sec:main-more-discussions} and Appendix~\ref{suppl:sec:more-discussion}.
More details are put in Appendix~\ref{suppl:sec:more-implementation}.

\begin{table}[t]
    % \vspace{-0.1in}
    \caption{Comparisons with SOTA methods on ImageNet-C (severity level 5) regarding \textbf{Accuracy~(\%)} under \texttt{continuous adaptation}. \textbf{BP} is short for backward propagation, \textbf{FTTA} is short for source-free fully TTA, and \textbf{\#FP} denotes the number of forward passes a sample. $\dagger$ indicates a variant that uses only entropy as the test-time objective, \ie, $\lambda{=}0$.
    The \textbf{bold} number indicates the best result.}
    \vspace{-0.1in}
    \setlength{\tabcolsep}{3.2pt}
    \label{tab:continuous-tto}
\newcommand{\tabincell}[2]{\begin{tabular}{@{}#1@{}}#2\end{tabular}}
 \begin{center}
 \begin{threeparttable}
 \LARGE
    \resizebox{1.0\linewidth}{!}{
\begin{tabular}{clcccccccccccccccccc>{\columncolor{blue!8}}c}
    \multicolumn{5}{c}{} 
    & \multicolumn{3}{c}{Noise} 
    & \multicolumn{4}{c}{Blur} 
    & \multicolumn{4}{c}{Weather} 
    & \multicolumn{4}{c}{Digital} 
    & \multicolumn{1}{c}{} \\

    \multicolumn{2}{l}{Model+Method} & BP & FTTA  & \#FP 
    & Gaus. & Shot & Imp. 
    & Def. & Glass & Mot. & Zoom 
    & Snow & Frost & Fog & Brit. 
    & Contr. & Elas. & Pix. & JPEG & Avg. \\

    \midrule

    \multirow{11}{*}{\rotatebox[origin=c]{90}{ViT-Base}}
    & NoAdapt & \ding{55} & \ding{52} & 1 & 56.8 & 56.8 & 57.5 & 46.9 & 35.6 & 53.1 & 44.8 & 62.2 & 62.5 & 65.7 & 77.7 & 32.6 & 46.0 & 67.0 & 67.6 & 55.5 \\

    & TENT & \ding{52} & \ding{52} & 1 & 60.3 & 63.8 & 64.8 & 55.0 & 55.7 & 60.9 & 56.4 & 61.8 & 63.4 & 70.0 & 78.6 & 64.5 & 50.1 & 70.2 & 70.5 & 63.1 \\

    & CoTTA & \ding{52} & \ding{52} & 3or35 & 63.4 & 64.3 & 61.6 & 35.9 & 50.1 & 47.7 & 44.6 & 38.2 & 48.8 & 42.3 & 54.2 & 5.2 & 55.5 & 56.3 & 54.7 & 48.2 \\

    & SAR & \ding{52} & \ding{52} & [1, 2] & 59.2 & 61.3 & 61.6 & 54.6 & 55.4 & 58.8 & 56.1 & 61.1 & 62.2 & 64.9 & 76.8 & 60.3 & 58.3 & 68.7 & 69.2 & 61.9 \\

    & DeYO & \ding{52} & \ding{52} & [1, 2] & 59.8 & 61.4 & 61.1 & 49.8 & 55.0 & 55.9 & 18.2 & 61.0 & 61.1 & 65.2 & 76.9 & 49.1 & 63.8 & 70.4 & 69.3 & 58.5 \\

    & ZOA & \ding{55} & \ding{55} & 2 & 58.2 & 60.5 & 61.7 & 49.9 & 46.9 & 57.6 & 54.4 & 64.3 & 62.8 & 66.3 & 78.5 & 59.6 & 55.7 & 69.5 & 70.6 & 61.1 \\

    & FOZO & \ding{55} & \ding{55} & 2 & 58.7 & 59.3 & 60.0 & 53.9 & 42.3 & 57.2 & 50.6 & 66.3 & 61.5 & 68.5 & 78.5 & 62.3 & 53.6 & 69.2 & 70.0 & 60.8  \\

    & FOA & \ding{55} & \ding{55} & 28 & 61.9 & 64.0 & 64.5 & 57.0 & 56.7 & 61.6 & 61.2 & 68.8 & 71.7 & 74.2 & 81.0 & 67.2 & 61.4 & 74.1 & 73.3 & 66.6 \\
    
    & FOA$^\dagger$ & \ding{55} & \ding{52} & 28 & 58.2 & 60.0 & 61.1 & 40.1 & 10.5 & 11.5 & 12.1 & 12.7 & 9.9 & 16.4 & 16.9 & 6.6 & 5.2 & 5.1 & 7.5 & 22.3 \\

    % \cmidrule(l){2-21}
    
     \rowcolor{black!4}\cellcolor{white} & \methodname (ours) & \ding{55} & \ding{55} & 2 & 61.8 & 63.6 & 64.2 & 58.3 & 56.3 & 62.3 & 60.9 & 70.1 & 69.2 & 74.1 & 81.0 & 68.9 & 66.6 & 74.0 & 73.3 & \textbf{67.0} \\

      \rowcolor{black!4}\cellcolor{white} & EVA-0$^\dagger$ (ours) & \ding{55} & \ding{52} & 2 & 60.1 & 62.0 & 62.6 & 55.8 & 53.3 & 61.2 & 57.8 & 67.3 & 66.0 & 70.9 & 79.6 & 66.8 & 62.0 & 72.1 & 72.1 & 64.6 \\

    \midrule

    \multirow{11}{*}{\rotatebox[origin=c]{90}{R50-GN}}
    & NoAdapt & \ding{55} & \ding{52} & 1 & 22.1 & 23.0 & 22.0 & 19.8 & 11.4 & 21.5 & 25.0 & 40.3 & 47.0 & 34.0 & 68.8 & 36.2 & 18.5 & 29.2 & 52.6 & 31.4 \\

    & TENT & \ding{52} & \ding{52} & 1 & 24.2 & 6.3 & 0.8 & 6.8 & 1.6 & 0.9 & 1.2 & 1.7 & 0.4 & 0.5 & 5.2 & 1.0 & 0.1 & 0.3 & 2.2 & 3.5 \\

    & CoTTA & \ding{52} & \ding{52} & 3or35 & 32.3 & 41.9 & 44.9 & 15.9 & 28.2 & 29.2 & 30.9 & 24.1 & 33.0 & 37.9 & 49.5 & 33.4 & 27.0 & 46.4 & 41.3 & 34.4 \\

    & SAR & \ding{52} & \ding{52} & [1, 2] & 32.1 & 38.7 & 38.7 & 27.7 & 30.5 & 35.7 & 42.9 & 36.7 & 36.7 & 47.7 & 59.3 & 32.2 & 47.5 & 50.8 & 47.5 & 40.3 \\

    & DeYO & \ding{52} & \ding{52} & [1, 2] & 40.7 & 48.4 & 47.9 & 9.2 & 12.1 & 16.3 & 18.5 & 3.1 & 0.8 & 0.3 & 0.8 & 0.2 & 0.1 & 0.1 & 0.1 & 13.2 \\

    % & T3A & \ding{55} & \ding{52} & 1 & 56.4 & 56.9 & 57.3 & 47.9 & 37.8 & 54.3 & 46.9 & 63.6 & 60.8 & 68.5 & 78.1 & 38.3 & 50.0 & 67.6 & 69.1 & 56.9 \\

    & ZOA & \ding{55} & \ding{55} & 2 & 29.0 & 27.6 & 27.8 & 24.2 & 14.0 & 25.0 & 33.3 & 42.0 & 42.3 & 53.8 & 70.0 & 43.7 & 26.8 & 34.4 & 54.1 & 36.5 \\

    & FOZO & \ding{55} & \ding{55} & 2 & 32.1 & 34.1 & 34.2 & 14.0 & 6.9 & 16.6 & 21.7 & 25.3 & 38.8 & 37.7 & 56.8 & 30.9 & 15.0 & 49.1 & 52.2 & 31.0  \\

    & FOA & \ding{55} & \ding{55} & 28 & 27.3 & 31.0 & 30.1 & 18.4 & 13.3 & 23.8 & 26.7 & 40.4 & 25.1 & 2.3 & 9.5 & 1.4 & 0.5 & 1.2 & 1.6 & 16.8 \\
    
    & FOA$^\dagger$ & \ding{55} & \ding{52} & 28 & 26.8 & 32.6 & 31.0 & 18.5 & 13.8 & 24.2 & 25.7 & 39.2 & 12.7 & 1.9 & 9.6 & 2.6 & 1.4 & 2.7 & 3.7 & 16.4 \\

    % \cmidrule(l){2-21}
    
     \rowcolor{black!4}\cellcolor{white} & \methodname (ours) & \ding{55} & \ding{55} & 2 & 37.0 & 40.4 & 39.5 & 28.2 & 23.1 & 38.4 & 45.1 & 50.3 & 51.8 & 60.2 & 72.3 & 49.1 & 40.5 & 52.6 & 56.0 & \textbf{45.6} \\

      \rowcolor{black!4}\cellcolor{white} & EVA-0$^\dagger$ (ours) & \ding{55} & \ding{52} & 2 & 32.3 & 38.4 & 37.5 & 24.3 & 18.8 & 33.5 & 38.2 & 45.3 & 47.8 & 53.0 & 71.4 & 47.3 & 29.3 & 50.6 & 56.6 & 41.6 \\      
      
\end{tabular}
	}
	 \end{threeparttable}
\vspace{-0.1in}	 
  \end{center}
\end{table}

\begin{table}[t]
    \caption{Efficacy of our \methodname on \textbf{Quantized ViT models}. We report \textbf{Accuracy (\%)} on ImageNet-C (level 5) under \texttt{single-domain TTA}.
    $\dagger$ indicates a variant that uses only entropy as the objective.
    }
    \vspace{-0.1in}
    \setlength{\tabcolsep}{3.6pt}
    \label{tab:quantized-vit}
\newcommand{\tabincell}[2]{\begin{tabular}{@{}#1@{}}#2\end{tabular}}
 \begin{center}
 \begin{threeparttable}
 \LARGE
    \resizebox{1.0\linewidth}{!}{
\begin{tabular}{clccccccccccccccccc>{\columncolor{blue!8}}c}
    \multicolumn{4}{c}{} 
    & \multicolumn{3}{c}{Noise} 
    & \multicolumn{4}{c}{Blur} 
    & \multicolumn{4}{c}{Weather} 
    & \multicolumn{4}{c}{Digital} 
    & \multicolumn{1}{c}{} \\

    \multicolumn{2}{l}{Model+Method} & FTTA  & \#FP 
    & Gaus. & Shot & Imp. 
    & Def. & Glass & Mot. & Zoom 
    & Snow & Frost & Fog & Brit. 
    & Contr. & Elas. & Pix. & JPEG & Avg. \\

    % \cmidrule(l){2-20}
    \midrule

    % \multirow{11}{*}{\rotatebox[origin=c]{90}{ViT-Base}}
    \multirow{7}{*}{8-bit}
    % \multirow{7}{*}{\rotatebox[origin=c]{90}{8-bit}}

    & NoAdapt & \ding{52} & 1 & 55.8 & 55.8 & 56.5 & 46.7 & 34.7 & 52.1 & 42.5 & 60.8 & 61.4 & 66.7 & 76.9 & 24.6 & 44.7 & 65.8 & 66.7 & 54.1 \\

    & ZOA & \ding{55} & 2 & 57.4 & 57.0 & 58.2 & 50.6 & 45.5 & 57.6 & 51.6 & 64.1 & 61.2 & 64.4 & 78.0 & 47.5 & 53.7 & 68.8 & 69.1 & 59.0 \\

    & FOZO & \ding{55} & 2 & 56.4 & 56.1 & 56.9 & 49.8 & 38.7 & 54.4 & 47.2 & 64.3 & 65.3 & 68.9 & 77.7 & 54.6 & 50.5 & 67.1 & 68.4 & 58.4  \\

    & FOA  & \ding{55} & 28 & 60.7 & 61.4 & 61.3 & 57.2 & 51.5 & 59.4 & 51.3 & 68.0 & 67.3 & 72.4 & 80.3 & 63.2 & 57.0 & 72.0 & 69.8 & 63.5 \\
    
    & FOA$^\dagger$ & \ding{52} & 28 &  57.1 & 57.7 & 58.2 & 52.1 & 35.6 & 54.5 & 35.3 & 46.0 & 48.6 & 38.3 & 76.7 & 21.5 & 47.6 & 65.5 & 66.6 & 50.8 \\

    % \cmidrule(l){2-21}
    
     \rowcolor{black!4}\cellcolor{white} & \methodname (ours)  & \ding{55} & 2 & 60.9 & 62.1 & 62.9 & 56.5 & 54.2 & 61.5 & 59.2 & 69.4 & 68.2 & 73.2 & 80.2 & 63.7 & 65.8 & 73.3 & 72.6 & \textbf{65.6} \\

      \rowcolor{black!4}\cellcolor{white} & EVA-0$^\dagger$ (ours) & \ding{52} & 2 & 59.1 & 60.4 & 61.0 & 56.2 & 53.3 & 60.4 & 56.8 & 67.4 & 66.4 & 71.4 & 79.4 & 62.2 & 62.6 & 71.9 & 71.7 & 64.0 \\ 
      \midrule

          \multirow{7}{*}{6-bit}
    % \multirow{7}{*}{\rotatebox[origin=c]{90}{8-bit}}

    & NoAdapt & \ding{52} & 1 & 44.2 & 42.0 & 44.8 & 39.8 & 28.9 & 43.4 & 34.7 & 53.2 & 59.8 & 59.0 & 75.1 & 27.4 & 39.0 & 59.1 & 65.3 & 47.7 \\

    & ZOA & \ding{55} & 2 & 47.5 & 45.5 & 46.8 & 41.2 & 37.5 & 47.9 & 41.9 & 55.6 & 60.7 & 60.4 & 75.6 & 14.8 & 49.0 & 62.3 & 66.5 & 50.2 \\

    & FOZO & \ding{55} & 2 & 41.9 & 43.4 & 44.6 & 41.1 & 34.8 & 46.1 & 40.3 & 56.6 & 61.2 & 64.0 & 75.1 & 26.6 & 47.2 & 59.7 & 67.0 & 50.0  \\

    & FOA  & \ding{55} & 28 & 53.2 & 51.8 & 54.6 & 49.6 & 38.8 & 51.0 & 44.8 & 60.3 & 65.0 & 68.8 & 76.7 & 39.5 & 46.6 & 67.3 & 68.6 & 55.8 \\
    
    & FOA$^\dagger$ & \ding{52} & 28 & 47.4 & 49.2 & 49.6 & 47.1 & 32.4 & 47.5 & 38.1 & 40.4 & 60.7 & 36.0 & 74.8 & 16.4 & 45.9 & 61.1 & 65.6 & 47.5 \\

    % \cmidrule(l){2-21}
    
     \rowcolor{black!4}\cellcolor{white} & \methodname (ours)  & \ding{55} & 2 & 54.1 & 55.4 & 56.0 & 51.0 & 48.8 & 55.9 & 53.7 & 63.7 & 65.4 & 69.3 & 78.1 & 45.6 & 61.3 & 69.0 & 70.8 & \textbf{59.9} \\

      \rowcolor{black!4}\cellcolor{white} & EVA-0$^\dagger$ (ours) & \ding{52} & 2 & 52.0 & 53.1 & 54.2 & 48.8 & 47.5 & 55.2 & 51.4 & 61.4 & 63.4 & 67.1 & 77.3 & 43.0 & 58.5 & 67.6 & 68.8 & 57.9 \\ 
      
\end{tabular}
	}
	 \end{threeparttable}
\vspace{-0.1in}	 
  \end{center}
\end{table}

\subsection{Main Results on Full Precision Models and Quantized Models}

\textbf{Single-domain test-time optimization} We first assess the effectiveness of \methodname in short-term TTO. Here, we conduct experiments under the single-domain TTA protocol, where the model is reset before evolving on each new target domain. From Table~\ref{tab:single-domain-tto}, we have the following observations. 
% 1) \methodname achieves the best average accuracy on both ViT-Base and ResNet50-GN under a strict two-forward budget without backpropagation, demonstrating its effectiveness. In particular, \methodname improves two-forward TTO through both informative direction estimation (Sec.~\ref{sec:method-direction}) and shortcut-resistant objective design (Sec.~\ref{sec:method-objective}), outperforming the strong BP-based method DeYO by 1.0\%, two-forward ZOA by 5.4\%, and 28-forward FOA by 0.4\% regarding the average accuracy on ViT-Base.
% 1) \methodname achieves the best average accuracy on both ViT-Base and ResNet50-GN under a strict two-forward budget without backpropagation. In particular, on ViT-Base, \methodname outperforms the strong BP-based method DeYO by 1.0\%, two-forward ZOA by 5.4\%, and 28-forward FOA by 0.4\% w.r.t. the average accuracy, suggesting its effectiveness.
1) \methodname achieves the best average accuracy on both ViT-Base and ResNet50-GN under a strict two-forward budget without backpropagation. On ViT-Base, \methodname outperforms the strong BP-based method DeYO by 1.0\%, two-forward ZOA by 5.4\%, and 28-forward FOA by 0.4\% w.r.t. the average accuracy, suggesting its effectiveness.
% 2) The advantage of \methodname becomes more pronounced on ResNet50-GN, outperforming FOA by 10.1\% while reducing forward evaluations by over 90\%. This suggests that \methodname is not tied to a specific architecture.
2) The advantage of \methodname becomes more pronounced on ResNet50-GN, outperforming FOA by 10.1\%, which suggests that \methodname is not tied to a specific architecture.
3) Importantly, \text{EVA-0}$^\dagger$ demonstrates the first practical \textit{forward-only fully TTA method}, which requires neither BP nor source statistics. On ViT-Base, EVA-0$^\dagger$ achieves 64.6\%, outperforming FOA$^\dagger$ (55.9\%) and even surpassing the BP-based SAR baseline (62.7\%).
These confirm the benefit of our informative direction estimation (Sec.~\ref{sec:method-direction}) and shortcut-resistant objective design (Sec.~\ref{sec:method-objective}).
% for test-time ZO update.

\textbf{Long-term continual test-time optimization} 
A practical TTO method should remain effective over online streams to continuously learn from evolving test-time experience.
We further conduct experiments under the continual TTA setting, where the model evolves across domains without reset, which stresses long-term TTO stability.
As shown in Table~\ref{tab:continuous-tto}, most existing BP-based and ZO-based methods significantly degrade in this setting, \ie, DeYO degrades from 65.7\% to 58.5\%, and FOA$^\dagger$ degrades from 55.9\% to 22.3\% w.r.t. accuracy on ViT-Base.
% This suggests effective short-term TTO does not necessarily translate into stable online evolution. 
In contrast, \methodname maintains and further improves performance, achieving the best average accuracy across models. 
Notably, \methodname and EVA-0$^\dagger$ achieve 45.6\% and 41.6\% average accuracy on ResNet50-GN, giving the best and second-best results.
This stability mainly benefits from anchor-guided optimization, as in Table~\ref{tab:eva0-ablation}, which allows \methodname to accumulate useful test-time signals without being dominated by noisy updates.

% \subsection{Results on Quantized Models}
% For efficient applications in practice, deep models often undergo quantization before deployment on edge devices, which renders BP infeasible for TTO. 
\textbf{Results on quantized models}
For practical edge deployment, deep models are often quantized before inference, which renders BP infeasible for TTO.
We further evaluate the effectiveness of \methodname on quantized models in Table~\ref{tab:quantized-vit}. \methodname significantly outperforms prior forward-only methods on both 8-bit and 6-bit ViT, whereas \methodname enables a 6-bit ViT to outperform a non-adapted 32-bit ViT by 4.4\% in OOD accuracy.
This suggests that \methodname offers a novel \textit{deployment tradeoff}: instead of relying on a fixed full-precision model for generalization, a quantized model can recover and even surpass full-precision robustness through lightweight forward-only evolution at testing.
Notably, even without using source feature statistics in a fully TTA manner, EVA-0$^\dagger$ outperforms FOA by 0.5\% on 8-bit ViT and 2.1\% on 6-bit ViT, with a \textit{14}-fold forward evaluation reduction. 
These results collectively underscore the superiority of our \methodname in such quantized model deployment scenarios.

\begin{table}[t]
    \centering
    \begin{minipage}[t]{0.41\textwidth}
        \caption{Ablation study on ResNet50-GN. $E^{\mathrm{SR}}$ and $\ell_{\mathrm{SWA}}$ are from Eqn.~(\ref{eq:overall-objective}). AGO is anchor-guided optimization, and SSD is sample-wise direction estimation. Single /Continual denotes settings in Tables~\ref{tab:single-domain-tto}/\ref{tab:continuous-tto}.}
        \vspace{-0.05in}
        \setlength{\tabcolsep}{7.3pt}
        % \vspace{-0.05in}
    	\label{tab:eva0-ablation}
        \newcommand{\tabincell}[2]{\begin{tabular}{@{}#1@{}}#2\end{tabular}}
        \begin{center}
        \begin{threeparttable}
            \resizebox{1.05\linewidth}{!}{
         	\begin{tabular}{cccc|cc}
                % \midrule
                $E^{\mathrm{SR}}$ & $\ell_{\mathrm{SWA}}$ & \textit{AGO} & \textit{SSD} & Single & Continual \\
                \midrule
                \multicolumn{4}{c|}{NoAdapt} & 31.4 & 31.4 \\
                \multicolumn{4}{c|}{ZOA} & 37.3 & 36.5 \\
                \ding{52} & & & & 39.5 & 29.2  \\
                & \ding{52} & & & 37.5 & 32.8 \\
                \ding{52} & \ding{52} & & & 41.3 & 35.4  \\
                \ding{52} & \ding{52} & \ding{52} &  & 41.1 & 42.2  \\
                \ding{52} & & & \ding{52} & 41.1 & 30.8 \\
                & \ding{52} & & \ding{52} & 42.6 & 40.1 \\
                \ding{52} & \ding{52} & & \ding{52} & 44.9 & 40.8 \\
                \midrule
                \ding{52} & \ding{52} & \ding{52} & \ding{52} & 45.4 & 45.6 \\

            \end{tabular}
            }
    	\end{threeparttable}
    	\end{center}
    	\vspace{0.02in}
    \end{minipage}
    ~~~~
    \begin{minipage}[t]{0.55\textwidth}
        \caption{
        Comparison of \textbf{throughput (FPS)} and \textbf{peak memory} on an NVIDIA Jetson Orin Nano with ViT-Base. Acc. (\%) is averaged under single-domain TTA on ImageNet-C (level 5). \#FP/\#BP: forward/backward passes per sample. OOM denotes out of memory.
        }
        \vspace{-0.05in}
        \setlength{\tabcolsep}{3.2pt}
    	\label{tab:computation-memory}
        \newcommand{\tabincell}[2]{\begin{tabular}{@{}#1@{}}#2\end{tabular}}
        \begin{center}
        \begin{threeparttable}
            \resizebox{0.95\linewidth}{!}{
         	\begin{tabular}{l|cccc|ccc}
                Method & BP & FTTA & \#FP & \#BP & Acc. & FPS & Mem. (MB) \\
                \midrule
                NoAdapt & \ding{55} & \ding{52} & 1 & 0 & 55.5 & 28.96 & 917 \\
                TENT & \ding{52} & \ding{52} & 1 & 1 & 59.6 & 14.03 & 3,425 \\
                CoTTA & \ding{52} & \ding{52} & 3or35 & 1 & 61.7 & OOM & OOM \\
                SAR & \ding{52} & \ding{52} & [1, 2] & [0, 2] & 62.7 & 7.02 & 3,425 \\
                DeYO & \ding{52} & \ding{52} & [1, 2] & [0, 1] & 65.7 & 10.05 & 3,186 \\
                ZOA & \ding{55} & \ding{55} & 2 & 0 & 61.3 & 14.70 & 918 \\
                FOZO & \ding{55} & \ding{55} & 2 & 0 & 61.7 & 14.58 & 923\\
                FOA & \ding{55} & \ding{55} & 28 & 0 & 66.3 & 1.00 & 924 \\
                \midrule
                \methodname (ours) & \ding{55} & \ding{55} & 2 & 0 & 66.7 & 14.08 & 919 \\
                EVA-0$^\dagger$(ours) & \ding{55} & \ding{52} & 2 & 0 & 64.6 & 14.08 & 919 \\
            \end{tabular}
            }
    	\end{threeparttable}
    	\end{center}
    	\vspace{0.02in}
    \end{minipage}
\end{table}

\begin{table}[t]
    \centering
    \begin{minipage}[t]{0.488\textwidth}
        \vspace{-0.02in}
        \caption{Efficacy on \textbf{black-box} model. \text{EVA-0$^{\hclubsuit}$} indicates a variant that uses only Eqn.~(\ref{eq:shortcut_resistant_entropy}) to update input prompts. We report Acc.(\%) over ImageNet-C (level 5) with ViT-B per Table~\ref{tab:single-domain-tto}.}
        \setlength{\tabcolsep}{2.3pt}
        % \vspace{-0.05in}
    	\label{tab:single_domain_TTA}
        \newcommand{\tabincell}[2]{\begin{tabular}{@{}#1@{}}#2\end{tabular}}
        \begin{center}
        \begin{threeparttable}
            \resizebox{0.98\linewidth}{!}{
         	\begin{tabular}{lccccc}
            Method & \#FP${=}2$ & \#FP${=}8$ & \#FP${=}14$ & \#FP${=}20$ & \#FP${=}26$ \\
            \midrule
            FOA$^\dagger$ & 58.8 & 57.6 & 55.9 & 55.0 & 55.9 \\
            EVA-0$^{\hclubsuit}$ (ours) & 64.1 & 65.2 & 65.7 & 65.8 & 66.1
            \end{tabular}
            }
    	\end{threeparttable}
    	\end{center}
    	\vspace{-0.1in}
    \end{minipage}
    ~~
    \begin{minipage}[t]{0.488\textwidth}
        \caption{
        Analysis on anchor design for \methodname. The anchor is updated: $\theta^{\mathrm{anc}}{\leftarrow} m\theta_{t}{+}(1{-}m) \theta^{\mathrm{anc}}$. $m=0$ corresponds to a fixed anchor. Single and Continual denote settings used in Tables~\ref{tab:single-domain-tto} \& \ref{tab:continuous-tto}.
        }
        \setlength{\tabcolsep}{5.75pt}
    	\label{tab:anchor-design}
        \newcommand{\tabincell}[2]{\begin{tabular}{@{}#1@{}}#2\end{tabular}}
        \begin{center}
        \begin{threeparttable}
            \resizebox{0.98\linewidth}{!}{
         	\begin{tabular}{lcccccc}
             & $m{=}1e$-$1$ & $1e$-$2$ & $1e$-$3$ & $1e$-$4$ & $1e$-$5$ & $0$\\
            \midrule
            Single & 45.6 & 45.5 & 45.5 & 45.4 & 45.4 & 45.4 \\
            Continual & 40.2 & 40.5 & 43.0 & 45.5 & 45.6 & 45.6 \\
            \end{tabular}
            }
    	\end{threeparttable}
    	\end{center}
    	\vspace{-0.1in}
    \end{minipage}
\end{table}

% \subsection{Ablation Studies}

% We ablate the key components of \methodname on ResNet50-GN in Table~\ref{tab:eva0-ablation}. First, shortcut-resistant entropy $E^{\mathrm{SR}}$ improves single-domain accuracy from 37.3\% (ZOA) to 39.5\%, verifying that it provides stronger feedback for ZO updates. Second, $\ell_{\mathrm{SWA}}$ provides a complementary signal, and combining it with $E^{\mathrm{SR}}$ improves single-domain accuracy to 41.3\%. Third, sample-wise direction estimation (SSD) better exploits the two-forward budget, improving $E^{\mathrm{SR}}+\ell_{\mathrm{SWA}}$ from 41.3\% to 44.9\% in the single-domain setting. Finally, anchor-guided optimization (AGO) is crucial for continual evolution: adding AGO improves the overall continual accuracy from 40.8\% to 45.6\%, and \methodname achieves the best results in both short-term and continual TTO with all components, verifying our effectiveness.

\subsection{More Discussions}\label{sec:main-more-discussions}

% \textbf{Effectiveness of components in \methodname} We ablate the key components of \methodname on ResNet50-GN in Table~\ref{tab:eva0-ablation}. First, $E^{\mathrm{SR}}$ improves single-domain accuracy from 37.3\% (ZOA) to 39.5\%, verifying that it provides stronger feedback for ZO updates. Second, $\ell_{\mathrm{SWA}}$ provides a complementary signal, and combining it with $E^{\mathrm{SR}}$ improves single-domain accuracy to 41.3\%. Third, sample-wise direction estimation (SSD) better exploits the two-forward budget, improving $E^{\mathrm{SR}}{+}\ell_{\mathrm{SWA}}$ from 41.3\% to 44.9\% in the single-domain setting. Finally, anchor-guided optimization (AGO) is crucial for continual evolution: adding AGO improves the continual accuracy from 40.8\% to 45.6\%, and \methodname achieves the best results in both short-term and continual TTO with all components, verifying our effectiveness.

\textbf{Effectiveness of components in \methodname} We ablate the key components of \methodname on ResNet50-GN in Table~\ref{tab:eva0-ablation}. First, $E^{\mathrm{SR}}$ improves single-domain accuracy from 37.3\% (ZOA) to 39.5\%, verifying that it provides stronger feedback for ZO updates. Second, $\ell_{\mathrm{SWA}}$ provides a complementary signal, and combining it with $E^{\mathrm{SR}}$ improves single-domain accuracy to 41.3\%. Third, sample-wise direction estimation (SSD) better exploits the two-forward budget, improving $E^{\mathrm{SR}}{+}\ell_{\mathrm{SWA}}$ from 41.3\% to 44.9\% in the single-domain setting. Finally, anchor-guided optimization (AGO) is crucial for continual evolution: adding AGO improves the continual accuracy from 40.8\% to 45.6\% by controlling the drift accumulated from random ZO updates, without which $E^{\mathrm{SR}}$ alone even falls below NoAdapt (29.2\% versus 31.4\%). With all components, \methodname achieves the best results in both settings.

% \textbf{Computation and memory efficiency}
% As shown in Table~\ref{tab:computation-memory}, \methodname achieves the best accuracy-efficiency trade-off. With only two forward passes and no backward pass, it obtains the highest accuracy of 66.7\% while using 241s runtime and 822MB memory on ViT-Base. Compared with BP-based DeYO, \methodname improves accuracy by 1.0 point, reduces runtime from 355s to 241s, and lowers memory usage from 5,350MB to 822MB. Compared with two-forward ZO baselines, \methodname keeps nearly the same runtime and memory as ZOA/FOZO, but improves the accuracy by about 5\%. Compared with FOA, \methodname achieves higher accuracy with $14\times$ fewer forward evaluations and about $15\times$ faster runtime. These results show that \methodname substantially improves the practicality of two-forward TTO by combining strong adaptation accuracy with low memory and computation cost.

\textbf{Computation and memory efficiency}
As shown in Table~\ref{tab:computation-memory}, \methodname achieves the best accuracy-efficiency trade-off on the edge device. With only two forward passes and no backward pass, it obtains the highest accuracy of 66.7\% at 14.08 FPS with 919MB memory. Compared with BP-based DeYO, \methodname improves accuracy by 1.0 point, raises throughput from 10.05 to 14.08 FPS, and lowers memory from 3,186MB to 919MB, while CoTTA even runs out of memory. Compared with ZOA/FOZO, \methodname keeps nearly the same throughput and memory but improves accuracy by about 5\%. Compared with FOA, \methodname achieves higher accuracy with $14\times$ fewer forward passes and $14\times$ higher throughput. Results on an A100 GPU show similar trends (Table~\ref{tab:computation-memory-a100} in Appendix~\ref{suppl:sec:more-discussion}). Overall, \methodname makes two-forward TTO practical with strong accuracy and low memory and compute cost.

\textbf{Efficacy of \methodname on black-box models}
We further evaluate \methodname in a black-box setting, where model weights are inaccessible and only input prompts are updated. As shown in Table~\ref{tab:single_domain_TTA}, FOA$^\dagger$ does not benefit from additional forward evaluations, degrading from 58.8\% at \#FP${=}2$ to around 55\% with larger budgets. In contrast, EVA-0$^{\hclubsuit}$, using only the shortcut-resistant entropy in Eq.~(\ref{eq:shortcut_resistant_entropy}), achieves 64.1\% with two forward passes and steadily improves to 66.1\% at \#FP${=}26$. This yields a 5.3\% gain under the strict two-forward constraint and a 10.2\% gain at \#FP${=}26$ over FOA$^\dagger$, showing that our objective provides stable source-free feedback that 
% enabling effective forward-only evolution even for black-box models.
enables effective black-box test-time evolution.

\begin{table}[t]
    \centering
    \newcommand{\tabincell}[2]{\begin{tabular}{@{}#1@{}}#2\end{tabular}}
    \begin{minipage}[t]{0.55\textwidth}
        \caption{\textbf{Generality} of our sample-wise perturbation and objective for other ZO methods. We apply them to CAZO~\cite{zhang2026curvature}, while keeping its learnable parameters and curvature-aware sampling on ImageNet-C \& ViT-B.}
        \vspace{-0.05in}
        \setlength{\tabcolsep}{6pt}
        \label{tab:cazo-combine}
        \begin{center}
            \resizebox{1.0\linewidth}{!}{%
            \begin{tabular}{lcc|cc}
                Method & \tabincell{c}{Sample-wise\\Perturbation} & \tabincell{c}{Our\\Objective} & \#FP & Acc. \\
                \midrule
                CAZO ($K$=1) & \ding{55} & \ding{55} & 3 & 61.9 \\
                CAZO ($K$=2) & \ding{55} & \ding{55} & 5 & 65.2 \\
                \rowcolor{black!4} CAZO ($K$=1) + ours & \ding{52} & \ding{52} & 3 & \textbf{67.3} \\
            \end{tabular}%
            }%
        \end{center}
    \end{minipage}
    ~~
    \begin{minipage}[t]{0.41\textwidth}
        \caption{Sensitivity to \textbf{batch size} $B$. We report the average accuracy (\%) on \mbox{ImageNet-C} (level 5) with \mbox{ResNet50-GN} under single-domain TTA.}
        \vspace{-0.05in}
        \setlength{\tabcolsep}{6pt}
        \label{tab:batch-size}
        \begin{center}
            \resizebox{0.92\linewidth}{!}{%
            \begin{tabular}{l|cccc}
                \multirow{2}{*}{Method} & \multicolumn{4}{c}{Batch size $B$} \\
                & 1 & 8 & 64 & 256 \\
                \midrule
                ZOA & 0.1 & 37.9 & 37.3 & 36.2 \\
                FOZO & 0.1 & 35.6 & 33.8 & 25.1 \\
                \rowcolor{black!4} \methodname (ours) & \textbf{45.4} & \textbf{45.5} & \textbf{45.4} & \textbf{45.2} \\
            \end{tabular}%
            }%
        \end{center}
    \end{minipage}
\end{table}

\textbf{Analysis on anchor design}
Table~\ref{tab:anchor-design} studies the anchor update rate $m$. In the single-domain setting, \methodname is robust to this choice, with all variants achieving around $45.4$--$45.6\%$ accuracy, where a moving anchor slightly improves the performance. In continual TTO, however, a fast-moving anchor significantly hurts stability, \ie, $m{=}10^{-1}$ and $10^{-2}$ only obtain $40.2\%$ and $40.5\%$, respectively. This indicates that if the anchor follows the online model too quickly, it absorbs noisy updates and loses its role as a stable reference. In contrast, a slowly moving or fixed anchor achieves the best continual accuracy, with $m=10^{-5}$ and $m=0$ both reaching $45.6\%$. Thus, a persistent anchor is important to avoid accumulated random drift. For simplicity, we use a fixed pretrained anchor in our \methodname.

\textbf{Generality of our designs to other ZO methods}
Our sample-wise perturbation and shortcut-resistant objective are not tied to the specific update rule of \methodname, but are \textit{general and broadly applicable}. To verify this, we plug them into CAZO~\cite{zhang2026curvature}, a recent ZO-TTA method that samples curvature-aware anisotropic perturbations with $K$ symmetric directions, while keeping its learnable parameters and perturbation sampling unchanged. As in Table~\ref{tab:cazo-combine}, with the same 3 forward passes ($K{=}1$, including inference), our designs significantly boost CAZO from 61.9\% to 67.3\%, even surpassing CAZO with $K{=}2$ (65.2\%) that requires 5 forward passes. This indicates that directional exploration and objective shortcuts are common bottlenecks of existing ZO-TTA methods rather than issues specific to our framework. Consequently, our designs are orthogonal to how perturbations are sampled and can be seamlessly plugged into other ZO-based TTA methods for better accuracy and query efficiency.

\textbf{Robustness to batch size}
We show that \methodname is insensitive to the batch size $B$. To keep the anchor-guided frequency consistent across batch sizes, our default $k{=}1$ is defined for $B{=}64$, \ie, one anchor-guided perturbation per 64 samples, and for other $B$ we maintain a sample counter to apply it every 64 samples. As shown in Table~\ref{tab:batch-size}, ZOA and FOZO collapse at $B{=}1$ (0.1\%), as their batch-level feature alignment becomes ill-defined with a single sample, and they also degrade as the batch grows, \eg, FOZO drops from 35.6\% at $B{=}8$ to 25.1\% at $B{=}256$, since a shared perturbation explores only one direction per batch. In contrast, \methodname remains stable at 45.2--45.5\% from $B{=}1$ to $B{=}256$. This is because sample-wise perturbations explore one direction per sample, thus the same samples yield equal direction explorations under any $B$, while EMA-based sample-wise feature alignment remains well-defined for a single sample. Thus, \methodname suits both streaming and large-batch deployment.

\section{Conclusion}
% \paragraph{Conclusion} 
In this paper, we establish that effective test-time model evolution is possible under an extremely restrictive yet deployment-realistic regime: only two forward passes per sample, without any backpropagation. We reveal that zeroth-order test-time optimization is fundamentally challenged by shortcut exploitation, accumulated weight drift, and less effective update direction estimation, and we address these challenges through EVA-0. By unifying shortcut-resistant objectives, anchor-guided trajectory control, and sample-wise symmetric two-sided perturbations, EVA-0 converts the minimal two-forward budget into both reliable inference and stable online learning. Across full-precision models, quantized networks, continual streams, and black-box settings, EVA-0 consistently delivers strong accuracy, low memory cost, and robust long-term stability. 
These findings demonstrate that forward-only model evolution is not merely a compromise for restricted scenarios, but a promising and scalable path toward adaptive deployed intelligence.
% These results show that forward-only test-time evolution can move beyond a constrained alternative to backpropagation, emerging as a practical and scalable paradigm for adaptive models in real-world deployment.

\section*{Acknowledgments}
This work is supported by a Start-up Grant from Nanyang Technological University, the Joint WeBank-NTU Research Institute on Fintech, Nanyang Technological University, Singapore, Jinan-NTU Green Technology Research Institute (GreenTRI), and the Centre of AI in Medicine (C-AIM), Nanyang Technological University, Singapore.

%%%%%%%%%%%%%%%%%%%%%%%%%%%%%%%%%%%%%%%%%%%%%%%%%%%%%%%%%%%%

% \clearpage
% \bibliography{conference}
% \bibliographystyle{conference}
{\bibliographystyle{abbrv} % use natbib
    % \small
    \bibliography{conference}
}

\newpage
\appendix
\def\cL{{\cal L}}

\setcounter{table}{0}
\setcounter{figure}{0} 
\renewcommand{\thetable}{\Alph{table}}
\renewcommand{\thefigure}{\Alph{figure}}
\renewcommand\theHtable{Appendix.\thetable}
\renewcommand\theHfigure{Appendix.\thefigure}
\newtcolorbox[auto counter]{mybox}[1][]{
  title=My Box \thetcbcounter,
  label=mybox:\thetcbcounter,
  #1
}

\def\mytitle{
\methodname: Test-Time Model Evolution with Only Two \\ Forward Passes per Sample
}

\begin{center}
	{

        \Large{\textbf{Supplementary Materials for}} \\
        \Large{\textbf{``\mytitle''}}
	}
\end{center}

\etocdepthtag.toc{mtappendix}
\etocsettagdepth{mtchapter}{none}
\etocsettagdepth{mtappendix}{subsection}

{
    \hypersetup{linkcolor=black}
    \footnotesize\tableofcontents
}

\clearpage

\section{Pseudo-Code of \methodname}\label{appx:sec:algorithm}

In this appendix, we provide the pseudo-code of our \methodname method. 
As shown in Algorithm~\ref{alg:eva0}, for each online test mini-batch $\mathcal{B}_t$, 
we first compute the anchor direction $d_t$ from the current online model to the anchor model, 
and sample perturbations from an adaptive distribution $\mathcal{D}_t$ 
(refer to lines 3--6). 
For each test sample $x_i$, \methodname performs two symmetric forward passes with 
the positively and negatively perturbed models, 
$f_{\theta_t+\mu z_{t,i}}$ and $f_{\theta_t-\mu z_{t,i}}$ 
(refer to line 7). 
These two forward passes are used for both inference and adaptation: 
the prediction is obtained by averaging the two perturbed outputs (refer to line 8), while the update direction is estimated from their shortcut-resistant loss difference 
(refer to lines 9-10). 
Specifically, the shortcut-resistant objective $\ell_{\mathrm{sr}}$ suppresses degenerate entropy-reduction shortcuts, and the sample-wise two-sided difference provides an informative zeroth-order direction under the two-forward constraint. 
After aggregating the sample-wise directions over the mini-batch, we update the online model with a zeroth-order gradient step 
(refer to lines 11-12). 
Lastly, we apply a weak weight relaxation toward the anchor model to softly bound the online trajectory and mitigate accumulated weight drift 
(refer to line 13). 
The online states, including the output center and target moments, are updated by EMA for subsequent test batches 
(refer to line 14).

\begin{algorithm}[h]
\caption{\methodname: Two-Forward Test-Time Model Evolution}
\label{alg:eva0}
\KwIn{Test stream $\{\mathcal{B}_t\}_{t=1}^{T}$, model $f_{\theta_0}$, learning rate $\eta$, hyper-parameters $\gamma=0.999$, $k=1$, $\mu$.}
\KwOut{Predictions $\{\hat y_{t,i}\}$}

Initialize $\theta_1 \leftarrow \theta_0$, $\theta^{\mathrm{anc}}\leftarrow\theta_0$, and online states $c_1,m_1,q_1 = \emptyset$\;

\For{$t=1,\ldots,T$}{
    Receive $\mathcal{B}_t=\{x_i\}_{i=1}^{B}$ and compute $d_t=\theta^{\mathrm{anc}}-\theta_t$\;
    Select anchor-guided subset $\mathcal{I}_t$ with $|\mathcal{I}_t|=k$\;

    \For{$i=1,\ldots,B$}{
        Sample $z_{t,i}\sim\mathcal{D}_t$ using random or anchor-guided perturbation \tcp*[r]{Eqn.~\eqref{eq:anchor-distribution}}
        \vspace{2pt}
        $(o^+_{t,i},h^+_{t,i})=f_{\theta_t+\mu z_{t,i}}(x_i)$,
        $(o^-_{t,i},h^-_{t,i})=f_{\theta_t-\mu z_{t,i}}(x_i)$ \tcp*[r]{two forwards}
        \vspace{2pt}

        $\hat y_{t,i}=\frac{1}{2}(o^+_{t,i}+o^-_{t,i})$ \tcp*[r]{inference}
        \vspace{2pt}

        $\ell^+_{t,i}=\ell_{\mathrm{sr}}(o^+_{t,i},h^+_{t,i})$,
        $\ell^-_{t,i}=\ell_{\mathrm{sr}}(o^-_{t,i},h^-_{t,i})$ \tcp*[r]{shortcut-resistant loss, Eqn.~\eqref{eq:overall-objective}}
        \vspace{2pt}

        $g_{t,i}=\frac{\ell^+_{t,i}-\ell^-_{t,i}}{2\mu}z_{t,i}$ \tcp*[r]{two-sided ZO direction}
        \vspace{2pt}
    }

    $\widehat g_t^{\mathrm{EVA}}=\frac{1}{B}\sum_{i=1}^{B}g_{t,i}$\;
    \vspace{2pt}
    $\theta'_{t+1}=\theta_t-\eta\widehat g_t^{\mathrm{EVA}}$\;
    \vspace{2pt}
    $\theta_{t+1}=(1-\gamma)\theta'_{t+1}+\gamma\theta^{\mathrm{anc}}$ \tcp*[r]{weight relaxation, Eqn.~\eqref{eq:weak-restore}}
    \vspace{2pt}

    Update $c_{t+1},m_{t+1},q_{t+1}$ by EMA via $\bar{o}_t=(o_t^++o_t^-)/2$, $\mathbb{E}[h=(h_t^++h_t^-)/2]$, $\mathbb{E}[h^2=({h_t^+}^2+{h_t^-}^2)/2]$ using moving average factor of $0.9$\;

}
\end{algorithm}

\clearpage
\section{Related Work}

\textbf{Backpropagation-based TTA} 
TTT~\cite{sun2020test} first pioneers BP-based TTA. During the training phase, TTT jointly trains the network with an auxiliary self-supervised branch, instantiated as rotation prediction~\cite{gidaris2018unsupervised}, and updates the shared encoder via the self-supervised objective at test time. The following works~\cite{gandelsman2022tmae,liu2024continual,osowiechi2024nc,bartler2022mt3} extend TTT from two axis: 
1) Auxiliary task design: TTT-MAE~\cite{gandelsman2022tmae} and CTMAE~\cite{liu2024continual} replace rotation prediction with masked image reconstruction~\cite{he2022masked} and histogram-of-oriented-gradients reconstruction. NC-TTT~\cite{osowiechi2024nc} and TTTFlow~\cite{osowiechi2023tttflow} introduce a discriminator on source and OOD features to pull features back to the source at testing. 
2) Boosting correlations between the auxiliary and the main task: MT3~\cite{bartler2022mt3} and Doc-TTT~\cite{gu2025docttt} exploit a meta-learning approach (MAML~\cite{finn2017model}) to improve the gradient alignment between the auxiliary and main tasks. S4T~\cite{jeong2025synchronizing} further synchronizes multiple downstream tasks during TTT by predicting inter-task relations.

TTT requires altering the model training process and access to the source data, which hinders its applicability to privacy-sensitive scenarios and off-the-shelf models. To address this, Fully TTA methods update any given model via unsupervised learning objectives, such as entropy minimization~\cite{wang2021tent,lee2024deyo,niu2023sar}, prediction consistency maximization~\cite{zhang2021memo,wang2022cotta}, energy adaptation~\cite{yuan2024tea, choi2024adaptive}, pseudo labeling~\cite{liang2020we,goyal2022test}, \etc~
However, these TTA methods still rely on backpropagation, which introduces substantial memory overhead, and can be incompatible with resource-limited edge devices, quantized models, specialized accelerators, or black-box model access settings.

\textbf{Forward-only \& optimization-free TTA} Without using backpropagation and optimizing model parameters, these methods seek model adaptation alone without any online learning at all. Existing methods can be categorized into three paradigms: 1) Statistics calibration~\cite{you2021test}, which replaces the source training statistics with statistics computed from the test data. 2) Input purification, which denoises the test data back to the source distribution via diffusion models~\cite{gao2023back} or Fast Fourier Transform~\cite{termohlen2021continual}. 3) Direct output adjustment, which maintains online class prototypes to replace the original classifier head for prediction~\cite{iwasawa2021test}, or calibrates output according to estimated priors~\cite{boudiaf2022parameter}.
However, without updating parameters online, these methods suffer from limited learning capability and fail to effectively transform unlabeled test-time experiences into adaptive knowledge, often resulting in limited performance gain when handling out-of-distribution data.

\textbf{Forward-optimization TTA} replaces the gradient-based parameter updates with derivative-free optimization strategies that require only forward evaluations. Recently, FOA explores this direction using CMA-ES~\cite{hansen2016cma} to evolve input prompts, which online maintains the covariance of update directions to guide future sampling. However, this process requires many forward evaluations, \eg, $K=28$ for stable learning signals, and is limited to a low-dimensional parameter space due to the high memory cost of the covariance matrix. ZOA~\cite{deng2025zoa} instead adopts one-sided SPSA~\cite{spall2002multivariate} to adapt quantized models using only two forward passes per sample. However, it leaves a significant performance gap compared to FOA, as shown in Table~\ref{tab:single-domain-tto}, due to its coarse and noisy signal from one-sided SPSA.
To this end, we study \textit{test-time model evolution} under a strict two-forward constraint. We reveal the limitations of prior ZO methods from the objective design, direction estimation, and weight dynamics, and propose \methodname to substantially advance the effectiveness and stability of two-forward model evolution.

\textbf{Zeroth-order optimization (ZOO)} estimates update directions from function-value queries, typically via finite differences~\cite{liu2020primer}, and thus avoids backpropagation. This makes it memory-efficient and suitable for memory-resource-constrained, quantized, or black-box models. Prior works have established its theoretical foundations~\cite{flaxman2005online,ghadimi2013stochastic,duchi2015optimal}. However, ZOO often suffers from high-variance gradient estimates and slow convergence in deep, high-dimensional models. To improve its efficiency, existing studies have made several attempts, such as sampling~\cite{cheng2021convergence} and sparsity-aware updates~\cite{cai2022zeroth}, with recent efforts further scaling ZOO to large models~\cite{malladi2023fine}. 

However, existing ZOO methods~\cite{malladi2023fine,chendeepzero} are still mainly designed for offline supervised settings, where many forward evaluations are affordable. In contrast, we study ZOO in a stricter test-time deployment setting, where only two forward passes are available for each test sample to support both inference and online adaptation. This setting is highly challenging yet under-explored due to its stringent efficiency constraint and unsupervised nature. We analyze the fundamental limitations of ZOO in this two-forward test-time model evolution setting and derive key insights to guide the design of more effective and stable methods.

\clearpage
\section{More Design Details of \methodname}\label{suppl:sec:more-design}

\subsection{Computational- and Memory-Efficient Implementation of \methodname}
\label{suppl:sec:efficient-implementation}

EVA-0 uses sample-wise independent perturbations, which could be naively implemented by constructing a different perturbed model for each sample. This implementation is inefficient: it would either serialize the forward computation over samples, or materialize a large perturbation tensor of size $B\times|\theta|$. We avoid this overhead with a layer-wise seeded implementation that preserves batch parallelism for most computations and only applies sample-wise perturbations to the learnable layers.

\textbf{Parallel computation}
Let $\theta=(\phi,\psi)$, where $\phi$ denotes frozen parameters and $\psi$ denotes the adapted parameters. EVA-0 preserves standard batched inference for the frozen part:
\begin{equation}
    X = f_{\phi}(\mathcal{B}).
\end{equation}
It then applies sample-wise perturbations only to the adapted part:
\begin{equation}
    y_i^{\pm}
    =
    f_{\phi,\,\psi\pm\mu z_i}(x_i),
    \qquad i=1,\ldots,B.
\end{equation}
In implementation, this does not require running $B$ independent models. Since $\psi$ contains only a small subset of layers, such as LayerNorm parameters, we form batched perturbed weights for these layers and compute them in parallel. Thus, EVA-0 keeps the efficiency of standard mini-batch inference for frozen layers, while introducing sample-wise perturbations only on a small fraction of layers where test-time optimization is performed, which introduces a minimal computation cost.

\textbf{Seeded layer-wise perturbation}
A naive implementation of sample-wise SPSA would store the full perturbation matrix $Z\in\mathbb{R}^{B\times|\theta|}$ for the mini-batch. This can dominate memory when the number of optimized parameters grows. Inspired by MeZO~\cite{malladi2023fine}, \methodname instead stores only random seeds. For each adapted layer $\ell$, we deterministically regenerate $z_{\ell,i}$ from the global seed and the layer index, use it to compute the perturbed layer output, and immediately discard it.

Formally, let $s_t$ be the random seed at step $t$. For each adapted layer $\ell$, EVA-0 samples
\[
    Z_{\ell,t}
    =
    \mathrm{Rand}(s_t,\ell)
    \in
    \mathbb{R}^{B\times|\theta_\ell|},
\]
uses $Z_{\ell,t}$ to compute the two perturbed forward passes, and then releases it before moving to the next layer. During gradient construction, EVA-0 regenerates the same $Z_{\ell,t}$ using the stored seed and multiplies it by the corresponding finite-difference scalar feedback.

\textbf{Memory cost analysis}
With the above implementation, the additional memory of EVA-0 is proportional to the largest adapted layer, rather than the total number of adapted parameters:
\[
    O\!\left(B\max_{\ell}|\theta_\ell|\right)
    \quad \text{instead of} \quad
    O(B|\theta|).
\]
Moreover, since EVA-0 is forward-only, it does not store intermediate activations for backpropagation. As a result, EVA-0 keeps the memory footprint close to standard inference while still supporting sample-wise independent perturbations and two-sided SPSA updates. In practice, after each adapted layer computes its perturbed output, the sampled perturbation and temporary batched parameters are \textit{immediately deleted}, and the perturbation is later regenerated from the seed to compute the gradient.

\begin{table}[h]
    \caption{Comparison of wall-clock time and memory for processing 50,000 ImageNet-C images on an A100 GPU with ViT-Base. Both variants lead to the same performance.
    }
    \vspace{-0.1in}

    \label{tab:different-implementation}
\newcommand{\tabincell}[2]{\begin{tabular}{@{}#1@{}}#2\end{tabular}}
 \begin{center}
 \begin{threeparttable}
 \LARGE
    \resizebox{0.7\linewidth}{!}{
\begin{tabular}{l|cc}

    & Time (s) & Memory (MB) \\
    \midrule
    \textit{Memory-efficiency-oriented implementation} & 241 & 822 \\
    \textit{Computation-efficiency-oriented implementation} & 236  & 828 \\

\end{tabular}
	}
	 \end{threeparttable}

  \end{center}
\end{table}

Table~\ref{tab:computation-memory} in the main paper follows the implementation as proposed in Sec.~\ref{suppl:sec:efficient-implementation}. The main extra computation cost, beyond doubling the forward passes, comes from the resampling of perturbations for three times: $f_{\theta_t+\mu z_{t,i}}(x_i)$, $f_{\theta_t-\mu z_{t,i}}(x_i)$, and gradient computation. One can trade memory efficiency for computational efficiency by sampling perturbation once, storing the perturbation, and reusing it for each sample. As shown in Table~\ref{tab:different-implementation}, this leads to a reduction of 5 seconds in latency. In practice, one can flexibly choose the appropriate implementation according to the used scenarios.

\subsection{Balancing Inward and Outward Updates with Anchor-Guided Perturbation}
\label{app:inward-outward-balance}

Although anchor-guided perturbations provide recovery-aware directions, repeated ZO updates may still accumulate biased drift away from the anchor. Since outward movement is riskier in unsupervised online evolution, EVA-0 further balances updates that move toward and away from the anchor.

Let $\Delta_t=-\eta\hat{g}_t$ be the ZO update and 
$
e_t=(\theta^{\mathrm{anc}}-\theta_t)/(\|\theta^{\mathrm{anc}}-\theta_t\|_2+\epsilon)
$
be the unit direction toward the anchor. We use 
$
c_t=\langle \Delta_t,e_t\rangle
$
to measure whether the update moves inward ($c_t>0$) or outward ($c_t<0$). EVA-0 maintains EMA estimates of recent inward and outward magnitudes:
\begin{equation}
    a_t^{\mathrm{in}}
    =
    \beta a_{t-1}^{\mathrm{in}}
    +
    (1-\beta)\max(c_t,0),
    \qquad
    a_t^{\mathrm{out}}
    =
    \beta a_{t-1}^{\mathrm{out}}
    +
    (1-\beta)\max(-c_t,0).
\end{equation}
When the current update moves away from the anchor, EVA-0 rescales only its anchor-parallel outward component by
\begin{equation}
    \alpha_t
    =
    \min\left(1,\frac{a_t^{\mathrm{in}}}{a_t^{\mathrm{out}}+\epsilon}\right),
    \qquad
    \widetilde{\Delta}_t
    =
    \Delta_t + (\alpha_t-1)c_t e_t
    \quad \text{if } c_t<0 .
\end{equation}
For inward updates, we simply set $\widetilde{\Delta}_t=\Delta_t$.

This strategy does not prevent the model from moving away from the anchor. It only weakens outward movement when recent outward updates dominate inward corrections to ensure long-term stability.

\section{More Implementation Details}\label{suppl:sec:more-implementation}

\subsection{More Details on Datasets}

\begin{figure*}[h]
\centering\includegraphics[width=0.6\linewidth]{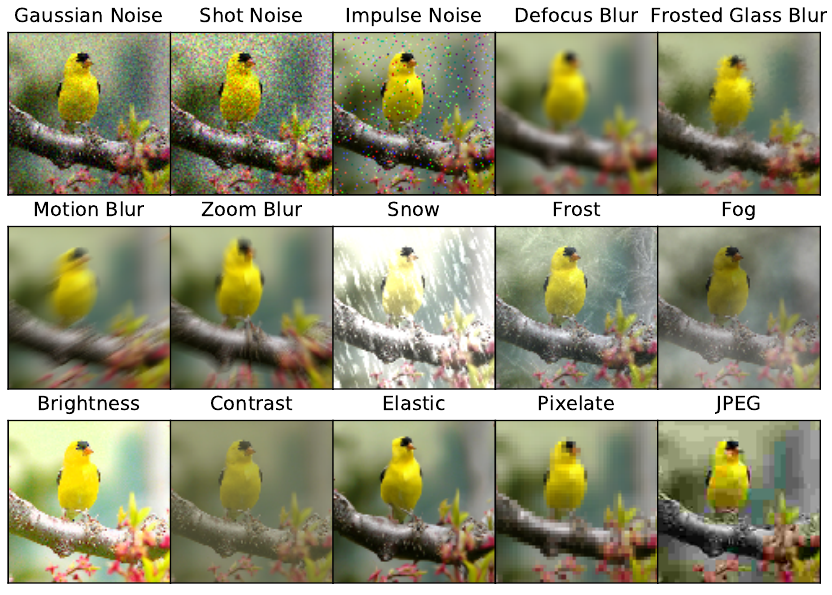}\caption{Visualizations of different corruption types in ImageNet-C benchmark.}
\label{fig:corruption_types}
\end{figure*}

We evaluate the test-time OOD generalization ability of all methods on a large-scale and widely used benchmark, namely \textbf{ImageNet-C}\footnote{\url{https://zenodo.org/record/2235448\#.YzQpq-xBxcA}}~\citep{hendrycks2019benchmarking}. ImageNet-C is constructed by corrupting the original ImageNet~\citep{deng2009imagenet} test set. The corruption (as shown in Figure~\ref{fig:corruption_types}) consists of 15 different types, \ie, Gaussian noise, shot noise, impulse noise, defocus blur, glass blur, motion blur, zoom blur, snow, frost, fog, brightness, contrast, elastic transformation, pixelation, and JPEG compression, where each corruption type has 5 severity levels and the larger severity level means more severe distribution shift. 

\subsection{More Experimental Protocols on Evaluation}

All pre-trained models involved in our paper are publicly available,
including ResNet50-GN\footnote{\url{https://github.com/rwightman/pytorch-image-models/releases/download/v0.1-rsb-weights/resnet50\_gn\_a1h2-8fe6c4d0.pth}} and ViT-Base\footnote{\url{https://storage.googleapis.com/vit\_models/augreg/B\_16-i21k-300ep-lr\_0.001-aug\_medium1-wd\_0.1-do\_0.0-sd\_0.0--imagenet2012-steps\_20k-lr\_0.01-res\_224.npz}} obtained from \texttt{timm} repository~\citep{rw2019timm}. In the following, we provide the implementation details of our proposed method and all comparative methods evaluated in our experiments, which help reproduce our results.

\textbf{\methodname (Ours).} We use source pretrained parameters as $\theta^{\mathrm{anc}}$ and set $\mu$ to 0.06 in Eqn.~(\ref{eq:eva0-gradient}), $k$ to 1 in Eqn.~(\ref{eq:anchor-distribution}), $\gamma$ to 0.001 in Eqn.~(\ref{eq:weak-restore}) and $\rho$ to 0.999 in Eqn.~(\ref{eq:smooth-for-align}), consistently for all experiments. We use SGD as the update rule, with a learning rate of 0.002 and a batch size of 64 for both models. For ViT-Base, we configure $\lambda$ to 500, and update online statistics $c_{t},m_{t},q_{t}$ using EMA with a factor of 0.9. For ResNet50-GN, we configure $\lambda$ to 50, and update online statistics $c_{t},m_{t},q_{t}$ using EMA with a factor of 0. Following ZOA, the trainable parameters are the affine parameters of norm layers from layer 1 to layer 3 in ResNet50-GN, from blocks 2 to blocks 8 in ViT-Base.

\textbf{TENT\footnote{\url{https://github.com/DequanWang/tent}}}~\citep{wang2021tent}. We follow all hyper-parameters that are set in TENT unless it does not provide. We use SGD as the update rule, with a momentum of 0.9, batch size of 64, and learning rate of 0.00025 / 0.001 for ResNet50-GN / ViT-Base. The trainable parameters are all the affine parameters of normalization layers for both models. 

\textbf{SAR\footnote{\url{https://github.com/mr-eggplant/SAR}}}~\citep{niu2023sar}. We follow all hyper-parameters that are set in SAR unless it does not provide. We use SGD as the update rule, with a momentum of 0.9, batch size of 64, and learning rate of 0.00025 / 0.001 for ResNet50-GN / ViT-Base. The entropy threshold $E_0$ is set to 0.4$\times\ln{C}$, where $C$ is the number of task classes. The trainable parameters are the affine parameters of norm layers from layer 1 to layer 3 in ResNet50-GN, from blocks 1 to blocks 8 in ViT-Base. 

\textbf{DeYO\footnote{\url{https://github.com/Jhyun17/DeYO}}}~\citep{lee2024deyo}. We follow all hyper-parameters that are set in DeYO unless it does not provide. Specifically, the entropy constant $E_0$ (for reliable sample identification) is set to $0.4\times\ln 1000$, and the factor $\tau_{\text{Ent}}$ is set to 0.5$\times\ln{1000}$. The Pseudo-Label Probability Difference (PLPD) threshold $\tau_{\text{PLPD}}$ is set to 0.2. The update rule is SGD, with a momentum of 0.9, batch size of 64, and learning rate of 0.00025 / 0.001 for ResNet50-GN / ViT-Base. Trainable parameters are the affine parameters of norm layers from layer 1 to 3 in ResNet50-GN, from blocks 1 to 8 in ViT-Base. 

\textbf{FOA\footnote{\url{https://github.com/mr-eggplant/FOA}}}~\citep{niu2024foa}. We follow all hyper-parameters that are set in FOA unless it does not provide. 
On ViT-Base, FOA adapts the model by learning 3 new input prompts, initialized uniformly. The trade-off parameter $\lambda$ in the fitness function is set to 0.4.
On ResNet50-GN, FOA updates an $7\times7$ convolutions applied before model inputs, initialized with zeros. The trade-off parameter $\lambda$ in the fitness function is set to $1/64$.
Optimization is performed via the derivative-free CMA-ES algorithm with a population size of 28. The trade-off parameter $\lambda$ in the fitness function is set to 0.4. For activation shifting, the step size $\gamma$ is 1.0, and historical statistics are updated with an exponential moving average factor $\alpha = 0.1$. The trainable parameters are the prompt embeddings.

\textbf{ZOA\footnote{\url{https://github.com/DengZeshuai/ZOA}}}~\citep{deng2025zoa}. We follow all hyper-parameters that are set in ZOA unless it does not provide. For the ViT model, perturbation vectors $\epsilon$ and $\nu$ are added with step sizes 0.02 and 0.05, respectively. The learning rates for parameters $\theta$ and $\alpha$ are 0.0005 and 0.01. The trainable parameters include the domain knowledge perturbation vectors, which are the affine parameters of the norm layers from layer 1 to 3 in ResNet50-GN, and from blocks 1 to 8 in ViT-Base.

\textbf{FOZO\footnote{\url{https://github.com/eVI-group-SCU/FOZO}}}~\citep{fozo2026}. We follow all hyper-parameters that are set in FOZO unless it does not provide. It employs an n-SPSA zero-order gradient estimator, where $n$ is set to 1 for 2 forward passes.
As in FOA, on ViT-Base, FOZO adapts the model by learning 3 new input prompts, initialized uniformly. The trade-off parameter $\lambda$ in the unsupervised loss is 0.4. 
On ResNet50-GN, updates an $7\times7$ convolutions applied before model inputs, initialized with zeros. The trade-off parameter $\lambda$ in the unsupervised loss is $1/64$.
The initial learning rate $\eta$ is 0.08. The decay factor $\alpha$ is 0.9, the threshold factor $\tau$ is 1.05, and the moving average factor $\beta$ for historical average loss is 0.9. 
Source training statistics are estimated using the validation set of ImageNet-1K. The trainable parameters are the prompt embeddings.

\clearpage

\section{Additional Discussions}\label{suppl:sec:more-discussion}

\textbf{Ablation on coefficient $\gamma$ in Eqn.~(\ref{eq:weak-restore}).}
We study the effect of the weak anchor relaxation in Table~\ref{tab:gamma-ablation}. After the ZO update produces an intermediate model $\theta_{t+1}^{\prime}$, EVA-0 applies Eqn.~(\ref{eq:weak-restore}) with relaxation coefficient $\gamma$, where a larger $\gamma$ indicates a stronger pull toward the anchor and $\gamma=0$ disables this relaxation. 
When the pull is too strong, \eg, $\gamma$=0.1, the model is overly constrained around the anchor and obtains only 39.1\% average accuracy. As $\gamma$ decreases, the model gains more plasticity and the performance improves, reaching the best accuracy of 45.6\% at $\gamma$=0.001. However, when the relaxation becomes too weak, \eg, $\gamma$=0.00001 or $\gamma$=0, the average accuracy drops to 43.5\%.

Compared with \methodname without AGO (37.0\%), anchor-guided optimization with the proper relaxation improves continual accuracy by 8.6\%. Compared with $\gamma$=0, the weak relaxation further brings a 2.1\% gain. These results confirm that a small but persistent restoring force is important for long-term TTO, whereas the anchor-guided perturbation also significantly improves the stability, 37.0\% $\rightarrow$ 43.5\%.

\begin{table}[h]

    \caption{Ablations of $\gamma$ on ImageNet-C (severity level 5) regarding \textbf{Accuracy~(\%)} under \texttt{continuous adaptation} with R50-GN. AGO is short for anchor-guided optimization. $\gamma$=0 disables Eqn.~(\ref{eq:weak-restore}).}
    \vspace{-0.1in}
    \setlength{\tabcolsep}{5pt}
    \label{tab:gamma-ablation}
\newcommand{\tabincell}[2]{\begin{tabular}{@{}#1@{}}#2\end{tabular}}
 \begin{center}
 \begin{threeparttable}
 \LARGE
    \resizebox{1.0\linewidth}{!}{
\begin{tabular}{lccccccccccccccc>{\columncolor{blue!8}}c}
    \multicolumn{1}{c}{} 
    & \multicolumn{3}{c}{Noise} 
    & \multicolumn{4}{c}{Blur} 
    & \multicolumn{4}{c}{Weather} 
    & \multicolumn{4}{c}{Digital} 
    & \multicolumn{1}{c}{} \\

    Method
    & Gaus. & Shot & Imp. 
    & Def. & Glass & Mot. & Zoom 
    & Snow & Frost & Fog & Brit. 
    & Contr. & Elas. & Pix. & JPEG & Avg. \\

    \midrule

    NoAdapt & 22.1 & 23.0 & 22.0 & 19.8 & 11.4 & 21.5 & 25.0 & 40.3 & 47.0 & 34.0 & 68.8 & 36.2 & 18.5 & 29.2 & 52.6 & 31.4 \\
    \methodname (w/o AGO) & 35.1 & 37.5 & 36.6 & 21.3 & 19.9 & 30.9 & 38.0 & 38.8 & 43.8 & 48.5 & 61.4 & 32.4 & 33.1 & 38.8 & 39.5 & 37.0 \\
    \methodname ($\gamma$=0.1) & 30.4 & 31.5 & 30.9 & 25.1 & 17.2 & 29.3 & 33.2 & 45.0 & 46.2 & 51.9 & 70.7 & 42.9 & 30.0 & 46.2 & 55.5 & 39.1  \\
    \methodname ($\gamma$=0.01) & 35.7 & 37.5 & 36.6 & 29.2 & 20.8 & 34.9 & 39.2 & 49.0 & 49.7 & 57.3 & 72.2 & 47.7 & 36.6 & 51.2 & 57.2 & 43.6 \\
    \methodname ($\gamma$=0.001) & 37.0 & 40.4 & 39.5 & 28.2 & 23.1 & 38.4 & 45.1 & 50.3 & 51.8 & 60.2 & 72.3 & 49.1 & 40.5 & 52.6 & 56.0 & 45.6 \\
    \methodname ($\gamma$=0.0001) & 36.6 & 39.3 & 37.9 & 25.0 & 24.1 & 35.4 & 44.2 & 49.0 & 51.6 & 57.9 & 71.1 & 45.8 & 41.3 & 50.3 & 51.5 & 44.1 \\
    \methodname ($\gamma$=0.00001) & 36.5 & 38.9 & 37.5 & 24.4 & 24.2 & 35.0 & 43.7 & 48.4 & 51.3 & 56.9 & 70.6 & 44.8 & 41.2 & 49.5 & 50.1 & 43.5 \\
    \methodname ($\gamma$=0) & 36.5 & 38.9 & 37.4 & 24.3 & 24.2 & 35.0 & 43.7 & 48.4 & 51.2 & 56.8 & 70.5 & 44.6 & 41.2 & 49.3 & 49.9 & 43.5\\
      
\end{tabular}
	}
	 \end{threeparttable}

  \end{center}
\end{table}

\textbf{Ablation on the number of anchor-guided perturbations $k$.}
We further ablate the number of anchor-guided perturbations $k$ in each mini-batch, as shown in Table~\ref{tab:k-ablation}. Without AGO, \methodname obtains only $37.0\%$ average accuracy under continual adaptation, showing that fully random ZO updates suffer from long-term drift. Introducing anchor-guided perturbations substantially improves stability: even $k$=1 improves the average accuracy to 45.6\%, bringing an 8.6\% gain over the variant without AGO. Increasing $k$ from $1$ to $5$ yields very similar performance, with accuracy remaining around $45.4\%$--$45.6\%$. This indicates that only a small amount of anchor-guided exploration is sufficient to provide a stable restoring direction. Therefore, we use $k$=1 by default, which provides strong stabilization while minimally interfering with random exploration.

\begin{table}[h]

    \caption{Ablations of $k$ on ImageNet-C (severity level 5) regarding \textbf{Accuracy~(\%)} under \texttt{continuous adaptation} with ResNet50-GN. AGO is short for anchor-guided optimization.}
    \vspace{-0.1in}
    \setlength{\tabcolsep}{5pt}
    \label{tab:k-ablation}
\newcommand{\tabincell}[2]{\begin{tabular}{@{}#1@{}}#2\end{tabular}}
 \begin{center}
 \begin{threeparttable}
 \LARGE
    \resizebox{1.0\linewidth}{!}{
\begin{tabular}{lccccccccccccccc>{\columncolor{blue!8}}c}
    \multicolumn{1}{c}{} 
    & \multicolumn{3}{c}{Noise} 
    & \multicolumn{4}{c}{Blur} 
    & \multicolumn{4}{c}{Weather} 
    & \multicolumn{4}{c}{Digital} 
    & \multicolumn{1}{c}{} \\

    Method
    & Gaus. & Shot & Imp. 
    & Def. & Glass & Mot. & Zoom 
    & Snow & Frost & Fog & Brit. 
    & Contr. & Elas. & Pix. & JPEG & Avg. \\

    \midrule

    NoAdapt & 22.1 & 23.0 & 22.0 & 19.8 & 11.4 & 21.5 & 25.0 & 40.3 & 47.0 & 34.0 & 68.8 & 36.2 & 18.5 & 29.2 & 52.6 & 31.4 \\
    \methodname (w/o AGO) & 35.1 & 37.5 & 36.6 & 21.3 & 19.9 & 30.9 & 38.0 & 38.8 & 43.8 & 48.5 & 61.4 & 32.4 & 33.1 & 38.8 & 39.5 & 37.0 \\
    \methodname ($k$=1) & 37.0 & 40.4 & 39.5 & 28.2 & 23.1 & 38.4 & 45.1 & 50.3 & 51.8 & 60.2 & 72.3 & 49.1 & 40.5 & 52.6 & 56.0 & 45.6 \\
    \methodname ($k$=2) & 36.8 & 40.3 & 39.3 & 27.9 & 23.3 & 37.3 & 44.3 & 50.3 & 52.0 & 59.8 & 72.3 & 48.3 & 41.0 & 53.6 & 56.6 & 45.5 \\
    \methodname ($k$=3) & 36.6 & 40.1 & 39.4 & 28.4 & 24.1 & 37.7 & 44.5 & 50.8 & 51.8 & 60.4 & 72.1 & 48.3 & 40.4 & 53.1 & 55.7 & 45.6 \\
    \methodname ($k$=4) & 36.3 & 40.3 & 39.3 & 27.5 & 22.9 & 38.0 & 44.4 & 49.7 & 52.1 & 60.0 & 72.2 & 48.4 & 41.0 & 53.0 & 56.4 & 45.4 \\
    \methodname ($k$=5) & 36.4 & 40.2 & 39.4 & 27.3 & 22.9 & 37.7 & 44.2 & 49.7 & 52.1 & 60.0 & 72.3 & 48.2 & 41.0 & 53.3 & 56.3 & 45.4 \\

\end{tabular}
	}
	 \end{threeparttable}

  \end{center}
\end{table}

\textbf{Effectiveness of $E^{\mathrm{SR}}$ under imbalanced target distributions.}
We further evaluate whether the proposed shortcut-resistant entropy remains effective under non-i.i.d. target streams with severe label imbalance. To isolate the effect of the objective, all methods in Table~\ref{tab:imbalanced-entropy} are optimized with backpropagation. Under the online imbalanced label-shift protocol of SAR~\cite{niu2023sar}, conventional entropy minimization becomes highly unstable: Tent drops to 23.8\% average accuracy, even worse than NoAdapt (31.4\%), and SAR only reaches 33.6\%. In contrast, our shortcut-resistant entropy in Eqn.~(\ref{eq:shortcut_resistant_entropy}) achieves 48.0\% average accuracy, improving over Tent, SAR, and DeYO by 24.2\%, 14.4\%, and 7.6\%, respectively. These results show that the benefit of $E^{\mathrm{SR}}$ is not limited to ZO optimization: by suppressing shortcut solutions such as confidence amplification and prediction collapse, it also provides a more reliable adaptation signal under challenging non-i.i.d. and imbalanced target distributions.

\begin{table}[h]

    \caption{Effectiveness of our shortcut-resistant entropy $E^{\mathrm{SR}}$ in Eqn.~(\ref{eq:shortcut_resistant_entropy}) under imbalanced target distributions.  Results are reported on ImageNet-C (severity level 5) under \textbf{\textsc{online imbalanced label shifts}} (imbalance ratio = $\infty$)  regarding \textbf{Accuracy (\%)} following SAR~\cite{niu2023sar}.}
    \vspace{-0.1in}
    \setlength{\tabcolsep}{5pt}
    \label{tab:imbalanced-entropy}
\newcommand{\tabincell}[2]{\begin{tabular}{@{}#1@{}}#2\end{tabular}}
 \begin{center}
 \begin{threeparttable}
 \LARGE
    \resizebox{1.0\linewidth}{!}{
\begin{tabular}{lccccccccccccccc>{\columncolor{blue!8}}c}
    \multicolumn{1}{c}{} 
    & \multicolumn{3}{c}{Noise} 
    & \multicolumn{4}{c}{Blur} 
    & \multicolumn{4}{c}{Weather} 
    & \multicolumn{4}{c}{Digital} 
    & \multicolumn{1}{c}{} \\

    Method
    & Gaus. & Shot & Imp. 
    & Def. & Glass & Mot. & Zoom 
    & Snow & Frost & Fog & Brit. 
    & Contr. & Elas. & Pix. & JPEG & Avg. \\

    \midrule

    NoAdapt & 22.1 & 23.0 & 22.0 & 19.8 & 11.4 & 21.5 & 25.0 & 40.3 & 47.0 & 34.0 & 68.8 & 36.2 & 18.5 & 29.2 & 52.6 & 31.4 \\
    TENT & 6.0 & 19.0 & 11.5 & 13.8 & 8.6 & 19.5 & 17.1 & 16.4 & 21.7 & 1.9 & 69.8 & 42.2 & 6.3 & 49.6 & 53.6 & 23.8 \\
    SAR & 36.5 & 38.1 & 37.0 & 19.4 & 18.1 & 33.4 & 16.1 & 22.6 & 44.4 & 3.8 & 71.6 & 46.6 & 7.1 & 52.0 & 56.5 & 33.6 \\
    DeYO & 43.4 & 44.9 & 43.3 & 22.7 & 24.0 & 40.5 & 5.0 & 52.3 & 51.0 & 0.8 & 73.3 & 52.4 & 19.0 & 58.8 & 59.0 & 39.4 \\
    \midrule
    $E^{\mathrm{SR}}$ (ours) & 38.2 & 41.2 & 40.0 & 28.8 & 28.0 & 40.2 & 44.7 & 56.6 & 52.0 & 60.9 & 72.3 & 51.2 & 48.7 & 59.5 & 57.7 & 48.0\\ 
    
\end{tabular}
	}
	 \end{threeparttable}

  \end{center}
\end{table}

\textbf{Efficiency comparison on an A100 GPU.}
In addition to the edge-device results in Table~\ref{tab:computation-memory}, we also compare the wall-clock time and memory of different methods on an A100 GPU in Table~\ref{tab:computation-memory-a100}, and observe consistent trends. With only two forward passes and no backward pass, \methodname obtains the highest accuracy of 66.7\% while using 241s runtime and 822MB memory on ViT-Base. Compared with BP-based DeYO, \methodname improves accuracy by 1.0 point, reduces runtime from 355s to 241s, and lowers memory usage from 5,350MB to 822MB. Compared with two-forward ZO baselines, \methodname keeps nearly the same runtime and memory as ZOA/FOZO, but improves the accuracy by about 5\%. Compared with FOA, \methodname achieves higher accuracy with $14\times$ fewer forward evaluations and about $15\times$ faster runtime. These results further confirm that \methodname combines strong adaptation accuracy with low memory and computation cost across different hardware platforms.

\begin{table}[h]
    \caption{Comparison of \textbf{wall-clock time} and \textbf{memory} for processing 50,000 ImageNet-C images on an A100 GPU with ViT-Base. Acc. (\%) is averaged under single-domain TTA on ImageNet-C (level 5). \#FP and \#BP are the number of forward and backward passes per sample.}
    \label{tab:computation-memory-a100}
    \vspace{-0.05in}
    \centering
    \setlength{\tabcolsep}{6pt}
    \resizebox{0.7\linewidth}{!}{
    \begin{tabular}{l|cccc|ccc}
        Method & BP & FTTA & \#FP & \#BP & Acc. & Time (s) & Mem. (MB) \\
        \midrule
        NoAdapt & \ding{55} & \ding{52} & 1 & 0 & 55.5 & 115 & 819 \\
        TENT & \ding{52} & \ding{52} & 1 & 1 & 59.6 & 245 & 5,165 \\
        CoTTA & \ding{52} & \ding{52} & 3or35 & 1 & 61.7 & 885 & 16,836 \\
        SAR & \ding{52} & \ding{52} & [1, 2] & [0, 2] & 62.7 & 488 & 5,166 \\
        DeYO & \ding{52} & \ding{52} & [1, 2] & [0, 1] & 65.7 & 355 & 5,350 \\
        ZOA & \ding{55} & \ding{55} & 2 & 0 & 61.3 & 234 & 822 \\
        FOZO & \ding{55} & \ding{55} & 2 & 0 & 61.7 & 248 & 830\\
        FOA & \ding{55} & \ding{55} & 28 & 0 & 66.3 & 3597 & 831 \\
        \midrule
        \methodname (ours) & \ding{55} & \ding{55} & 2 & 0 & 66.7 & 241 & 822 \\
        EVA-0$^\dagger$ (ours) & \ding{55} & \ding{52} & 2 & 0 & 64.6 & 241 & 822 \\
    \end{tabular}
    }
\end{table}

\textbf{Comparison with normalization-statistics adaptation.}
A simple alternative to learning-based TTA is to re-estimate normalization statistics on test data without updating any parameters, \eg, AdaBN~\cite{li2016revisiting} and SITA~\cite{khurana2021sita}. As shown in Table~\ref{tab:norm-stat}, on ResNet-50 (BN), AdaBN and SITA improve NoAdapt from 18.2\% to 31.7\% and 31.8--32.2\%, respectively, while \methodname further reaches 34.7\%, showing that feedback-driven optimization brings gains beyond statistics re-estimation. Notably, \methodname also requires fewer sample-equivalent evaluations than SITA (2 versus 3 or 5). More importantly, statistics adaptation applies only to models with batch normalization, since group and layer normalization do not maintain batch-dependent running statistics. On ResNet-50 (GN) and ViT-Base (LN), where AdaBN and SITA are inapplicable, \methodname still improves accuracy from 31.4\% to 45.4\% and from 55.5\% to 66.7\%, respectively, demonstrating its broad applicability to modern architectures.

\begin{table}[h]
    \caption{Comparison with normalization-statistics adaptation methods across architectures. We report average accuracy (\%) over 15 corruptions on ImageNet-C (severity level 5). \#FP counts sample-equivalent forward evaluations per sample, \eg, SITA with $k$ augmented views costs $k{+}1$. N/A: not applicable, since GN and LN maintain no batch-dependent running statistics.}
    \vspace{-0.05in}
    \label{tab:norm-stat}
    \centering
    \setlength{\tabcolsep}{8pt}
    \resizebox{0.75\linewidth}{!}{%
    \begin{tabular}{lc|ccc}
        Method & \#FP & ResNet-50 (BN) & ResNet-50 (GN) & ViT-Base (LN) \\
        \midrule
        NoAdapt & 1 & 18.2 & 31.4 & 55.5 \\
        AdaBN & 1 & 31.7 & N/A & N/A \\
        SITA ($k$=2) & 3 & 31.8 & N/A & N/A \\
        SITA ($k$=4) & 5 & 32.2 & N/A & N/A \\
        \midrule
        \methodname (ours) & 2 & \textbf{34.7} & \textbf{45.4} & \textbf{66.7} \\
    \end{tabular}%
    }
\end{table}

\textbf{Robustness under online imbalanced label shifts.}
Following SAR~\cite{niu2023sar}, we evaluate \methodname under online imbalanced label shifts (imbalance ratio $=\infty$), where the label distribution of the test stream changes abruptly over time. As shown in Table~\ref{tab:label-shift}, TENT collapses on several corruptions (\eg, 1.1\% on Fog), and the ZO baselines ZOA, FOZO, and FOA only reach 57.5--62.5\%. In contrast, \methodname achieves the best average accuracy of 66.0\%, outperforming FOA by 3.5\% and even the BP-based SAR and DeYO without backpropagation. This benefits from two designs: the decentering term in Eqn.~(\ref{eq:shortcut_resistant_entropy}) prevents predictions from collapsing to the currently dominant classes, while EMA-based statistics aggregate the whole stream and remain stable under abrupt label changes.

\begin{table}[h]
    \caption{Comparisons with state-of-the-art methods on ImageNet-C (severity level 5) under \textbf{\textsc{online imbalanced label shifts}} (imbalance ratio = $\infty$) following SAR~\cite{niu2023sar}, regarding \textbf{Accuracy (\%)} with ViT-Base. The \textbf{bold} number indicates the best result.}
    \vspace{-0.1in}
    \setlength{\tabcolsep}{5pt}
    \label{tab:label-shift}
\newcommand{\tabincell}[2]{\begin{tabular}{@{}#1@{}}#2\end{tabular}}
 \begin{center}
 \begin{threeparttable}
 \LARGE
    \resizebox{1.0\linewidth}{!}{
\begin{tabular}{lcccccccccccccccc>{\columncolor{blue!8}}c}
    \multicolumn{2}{c}{} 
    & \multicolumn{3}{c}{Noise} 
    & \multicolumn{4}{c}{Blur} 
    & \multicolumn{4}{c}{Weather} 
    & \multicolumn{4}{c}{Digital} 
    & \multicolumn{1}{c}{} \\

    Method & BP
    & Gaus. & Shot & Imp. 
    & Def. & Glass & Mot. & Zoom 
    & Snow & Frost & Fog & Brit. 
    & Contr. & Elas. & Pix. & JPEG & Avg. \\

    \midrule

    NoAdapt & \ding{55} & 56.8 & 56.8 & 57.5 & 46.9 & 35.6 & 53.1 & 44.8 & 62.2 & 62.5 & 65.7 & 77.7 & 32.6 & 46.0 & 67.0 & 67.6 & 55.5 \\
    TENT & \ding{52} & 61.3 & 62.5 & 62.7 & 59.7 & 58.5 & 64.7 & 60.9 & 28.9 & 67.0 & 1.1 & 79.7 & 67.1 & 65.4 & 74.1 & 71.6 & 59.0 \\
    SAR & \ding{52} & 59.6 & 60.7 & 60.8 & 57.8 & 57.6 & 62.6 & 59.6 & 66.4 & 64.8 & 68.3 & 78.7 & 64.5 & 65.3 & 73.1 & 70.8 & 64.7 \\
    DeYO & \ding{52} & 58.4 & 59.5 & 59.2 & 54.7 & 56.5 & 62.9 & 61.2 & 68.2 & 65.8 & 68.9 & 77.8 & 59.9 & 69.7 & 73.7 & 71.6 & 64.5 \\
    ZOA & \ding{55} & 56.4 & 57.1 & 57.8 & 50.9 & 47.5 & 57.1 & 51.4 & 63.8 & 60.1 & 64.4 & 77.4 & 53.1 & 56.4 & 67.5 & 67.8 & 59.2 \\
    FOZO & \ding{55} & 55.8 & 56.2 & 55.3 & 50.1 & 39.6 & 53.5 & 47.5 & 63.6 & 60.6 & 64.7 & 75.9 & 56.2 & 50.1 & 66.7 & 67.1 & 57.5 \\
    FOA & \ding{55} & 58.5 & 59.9 & 60.2 & 56.2 & 50.1 & 57.7 & 53.3 & 66.2 & 66.7 & 69.3 & 77.8 & 63.1 & 58.5 & 69.7 & 70.9 & 62.5 \\
    \midrule
    \methodname (ours) & \ding{55} & 60.7 & 62.0 & 62.6 & 57.4 & 55.5 & 61.7 & 60.3 & 69.3 & 67.8 & 73.1 & 79.6 & 68.0 & 66.7 & 72.8 & 71.8 & \textbf{66.0} \\
\end{tabular}
	}
	 \end{threeparttable}
  \end{center}
\end{table}

\textbf{Robustness under natural domain shifts.}
Beyond synthetic corruptions, we evaluate \methodname on DomainNet~\cite{peng2019moment} with ResNet-50 (GN) under continual TTA, where each domain serves as the source and the model adapts sequentially to the other three domains. As shown in Table~\ref{tab:domainnet}, \methodname achieves the best accuracy under every source domain and improves NoAdapt from 59.2\% to 62.6\% on average, outperforming the BP-free ZOA (60.6\%) and even the BP-based TENT, SAR, and DeYO (59.4--61.4\%) with only two forward passes. These results show that \methodname generalizes from synthetic corruptions to natural domain shifts and remains stable under continual domain changes.

\begin{table}[h]
    \caption{Comparisons on \textbf{DomainNet} with ResNet-50 (GN) under \texttt{continuous adaptation} regarding \textbf{Accuracy (\%)}. Each domain serves as the source, and the model adapts sequentially to the other three domains in the listed order. The \textbf{bold} number indicates the best result.}
    \vspace{-0.1in}
    \setlength{\tabcolsep}{5pt}
    \label{tab:domainnet}
 \begin{center}
 \begin{threeparttable}
 \LARGE
    \resizebox{1.0\linewidth}{!}{
\begin{tabular}{lc|cccc|cccc|cccc|cccc|>{\columncolor{blue!8}}c}
    \multicolumn{2}{c}{Source:}
    & \multicolumn{4}{c}{Real}
    & \multicolumn{4}{c}{Painting}
    & \multicolumn{4}{c}{Clipart}
    & \multicolumn{4}{c}{Sketch}
    & \multicolumn{1}{c}{Overall} \\
    Method & BP
    & Clp. & Pnt. & Skt. & Avg.
    & Rel. & Skt. & Clp. & Avg.
    & Skt. & Rel. & Pnt. & Avg.
    & Pnt. & Clp. & Rel. & Avg. & Avg. \\
    \midrule
    NoAdapt & \ding{55} & 55.0 & 65.3 & 47.8 & 56.0 & 78.4 & 51.2 & 54.3 & 61.3 & 52.0 & 66.2 & 51.4 & 56.5 & 59.3 & 60.7 & 68.7 & 62.9 & 59.2 \\
    TENT & \ding{52} & 55.1 & 65.7 & 49.5 & 56.8 & 78.6 & 51.0 & 55.0 & 61.6 & 52.1 & 66.3 & 52.1 & 56.9 & 59.0 & 60.4 & 67.3 & 62.2 & 59.4 \\
    SAR & \ding{52} & 55.1 & 65.7 & 49.5 & 56.8 & 78.5 & 51.5 & 55.0 & 61.7 & 52.3 & 66.4 & 52.4 & 57.0 & 59.6 & 61.2 & 69.0 & 63.2 & 59.7 \\
    DeYO & \ding{52} & 56.3 & 66.9 & 53.6 & 58.9 & 78.9 & 55.1 & 58.1 & 64.0 & 52.9 & 67.0 & 54.8 & 58.3 & 60.6 & 62.6 & 70.2 & 64.5 & 61.4 \\
    ZOA & \ding{55} & 56.2 & 65.8 & 50.3 & 57.4 & 78.4 & 55.3 & 55.4 & 63.0 & 54.5 & 65.6 & 52.4 & 57.5 & 62.3 & 61.7 & 69.0 & 64.4 & 60.6 \\
    \midrule
    \methodname (ours) & \ding{55} & 58.4 & 66.5 & 53.4 & \textbf{59.4} & 78.5 & 58.7 & 59.0 & \textbf{65.4} & 56.0 & 67.3 & 54.7 & \textbf{59.3} & 64.1 & 64.1 & 70.9 & \textbf{66.3} & \textbf{62.6} \\
\end{tabular}
	}
	 \end{threeparttable}
  \end{center}
\end{table}

\textbf{Sensitivity to perturbation scale $\mu$.}
We study the effect of the perturbation scale $\mu$ in Table~\ref{tab:mu-sensitivity}. \methodname is robust to $\mu$ over a wide range: for $\mu\in[0.01, 0.1]$, its accuracy varies within 0.5\% on ViT-Base (66.2--66.7\%) and within 1.4\% on ResNet50-GN (44.1--45.5\%). This is consistent with Figure~\ref{fig:motivation_3}~(a), where symmetric two-sided probing remains stable across perturbation scales, while one-sided SPSA is more sensitive. Notably, we simply set $\mu{=}0.06$ by default without tuning. When $\mu$ exceeds 0.1, the performance degrades, \eg, to 59.5\% and 31.6\% at $\mu{=}0.2$, since a larger perturbation increases the Taylor approximation error, which negatively affects both the update direction estimation and the averaged prediction for inference (\cf~Sec.~\ref{sec:direction_motivation}).

\begin{table}[h]
    \caption{Sensitivity of the perturbation scale $\mu$ on ImageNet-C (severity level 5) regarding \textbf{Accuracy (\%)} under \texttt{single-domain TTA}. We use $\mu{=}0.06$ by default.}
    \vspace{-0.1in}
    \setlength{\tabcolsep}{5pt}
    \label{tab:mu-sensitivity}
 \begin{center}
 \begin{threeparttable}
    \resizebox{0.8\linewidth}{!}{
\begin{tabular}{l|ccccc>{\columncolor{blue!8}}ccccccc}
    $\mu$ & 0.01 & 0.02 & 0.03 & 0.04 & 0.05 & 0.06 & 0.07 & 0.08 & 0.09 & 0.10 & 0.15 & 0.20 \\
    \midrule
    ViT-Base & 66.5 & 66.5 & 66.6 & 66.7 & 66.7 & 66.7 & 66.7 & 66.6 & 66.4 & 66.2 & 64.1 & 59.5 \\
    ResNet50-GN & 44.8 & 45.0 & 45.2 & 45.4 & 45.5 & 45.4 & 45.3 & 45.0 & 44.6 & 44.1 & 39.3 & 31.6 \\
\end{tabular}
	}
	 \end{threeparttable}
  \end{center}
\end{table}

\textbf{Sensitivity to alignment coefficient $\lambda$.}
We study the effect of the coefficient $\lambda$ of $\ell_{\mathrm{SWA}}$ in Table~\ref{tab:lambda-sensitivity}, where $\lambda_0$ denotes our default value ($\lambda_0{=}500$ for ViT-Base and $\lambda_0{=}50$ for ResNet50-GN). We choose $\lambda_0$ such that the entropy and alignment terms produce gradients of comparable magnitudes, which requires no labels. \methodname is robust to $\lambda$ over a wide range, achieving over 66.0\% on ViT-Base for $\lambda\in[0.4\lambda_0, 2\lambda_0]$ and at least 45.0\% on ResNet50-GN for $\lambda\in[0.6\lambda_0, 2\lambda_0]$. As $\lambda$ decreases further, the performance gradually drops toward the entropy-only variant EVA-0$^\dagger$ ($\lambda{=}0$), indicating that sample-wise feature alignment provides a complementary signal to the shortcut-resistant entropy.

\begin{table}[h]
    \caption{Sensitivity of the alignment coefficient $\lambda$ on ImageNet-C (severity level 5) regarding \textbf{Accuracy (\%)} under \texttt{single-domain TTA}. $\lambda_0{=}500$ for ViT-Base and $\lambda_0{=}50$ for ResNet50-GN. The default $\lambda_0$ is highlighted.}
    \vspace{-0.1in}
    \setlength{\tabcolsep}{5pt}
    \label{tab:lambda-sensitivity}
 \begin{center}
 \begin{threeparttable}
    \resizebox{0.85\linewidth}{!}{
\begin{tabular}{l|ccccc>{\columncolor{blue!8}}cccccc}
    $\lambda$ & 0.1$\lambda_0$ & 0.2$\lambda_0$ & 0.4$\lambda_0$ & 0.6$\lambda_0$ & 0.8$\lambda_0$ & $\lambda_0$ & 1.2$\lambda_0$ & 1.4$\lambda_0$ & 1.6$\lambda_0$ & 1.8$\lambda_0$ & 2.0$\lambda_0$ \\
    \midrule
    ViT-Base & 64.7 & 65.4 & 66.1 & 66.5 & 66.7 & 66.7 & 66.7 & 66.7 & 66.6 & 66.5 & 66.4 \\
    ResNet50-GN & 42.4 & 43.4 & 44.5 & 45.0 & 45.3 & 45.4 & 45.5 & 45.5 & 45.5 & 45.5 & 45.5 \\
\end{tabular}
	}
	 \end{threeparttable}
  \end{center}
\end{table}

\textbf{Effect of output decentering in $E^{\mathrm{SR}}$.}
Eqn.~(\ref{eq:shortcut_resistant_entropy}) combines scale normalization with output decentering, which address two different shortcuts of entropy minimization. Scale normalization prevents reducing entropy by inflating the logit norm, yet entropy can still be minimized by collapsing predictions into a few dominant classes, \eg, confidently predicting ``cat'' for all inputs. Decentering subtracts the online output center $c_t$, which captures the persistent output bias from historical predictions, so that this bias is not further reinforced during adaptation. To isolate its effect, we optimize all variants with BP under the online imbalanced label shift protocol of SAR~\cite{niu2023sar}, where such collapse is most likely. As shown in Table~\ref{tab:decentering}, making entropy scale-invariant improves TENT from 23.8\% to 31.7\%, and further subtracting $c_t$ boosts the accuracy to 48.0\%, showing that both terms are necessary for shortcut-resistant entropy minimization.

\begin{table}[h]
    \caption{Effect of output decentering in Eqn.~(\ref{eq:shortcut_resistant_entropy}). All methods are optimized with BP for a direct comparison. We report average accuracy (\%) over 15 corruptions on ImageNet-C (severity level 5) with ResNet50-GN under online imbalanced label shifts (imbalance ratio $=\infty$).}
    \vspace{-0.05in}
    \label{tab:decentering}
    \centering
    \setlength{\tabcolsep}{9pt}
    \resizebox{0.6\linewidth}{!}{%
    \begin{tabular}{lcc|c}
        Method & Scale norm. & Decentering & Acc. \\
        \midrule
        NoAdapt & -- & -- & 31.4 \\
        SAR & -- & -- & 33.6 \\
        DeYO & -- & -- & 39.4 \\
        \midrule
        Entropy (TENT) & \ding{55} & \ding{55} & 23.8 \\
        $E^{\mathrm{SR}}$ w/o decentering & \ding{52} & \ding{55} & 31.7 \\
        $E^{\mathrm{SR}}$ (ours) & \ding{52} & \ding{52} & \textbf{48.0} \\
    \end{tabular}%
    }
\end{table}

\textbf{Sensitivity to random seeds and corruption orders.}
We evaluate the sensitivity of \methodname to perturbation seeds under single-domain TTA and to corruption orders under continual TTA in Table~\ref{tab:seed-order}. \methodname consistently achieves the highest accuracy with the lowest standard deviation in both settings. For example, under continual TTA, \methodname improves the accuracy of ZOA from 35.1\% to 45.6\% while reducing the standard deviation from 1.01 to 0.04, \ie, by over 20$\times$. In contrast, BP-based TENT and SAR and the evolution-based FOA vary substantially across corruption orders, \eg, FOA has a standard deviation of 7.44. This insensitivity to perturbation seeds and corruption orders further underscores the reliability of \methodname for real-world deployment.

\begin{table}[h]
    \caption{Sensitivity to random seeds and corruption orders on ImageNet-C (severity level 5) with ResNet50-GN regarding \textbf{Accuracy (\%)}. We report the mean and standard deviation over 3 random seeds for \texttt{single-domain TTA} and 3 different corruption orders for \texttt{continual TTA}, following the protocols of Tables~\ref{tab:single-domain-tto} and~\ref{tab:continuous-tto}.}
    \vspace{-0.05in}
    \label{tab:seed-order}
    \centering
    \setlength{\tabcolsep}{6pt}
    \resizebox{0.9\linewidth}{!}{%
    \begin{tabular}{l|ccccc>{\columncolor{blue!8}}c}
        Setting & TENT & SAR & ZOA & FOZO & FOA & \methodname (ours) \\
        \midrule
        Single-domain (3 seeds) & 23.4\scriptsize{$\pm$3.96} & 37.5\scriptsize{$\pm$2.21} & 37.1\scriptsize{$\pm$0.09} & 33.7\scriptsize{$\pm$0.17} & 35.3\scriptsize{$\pm$0.09} & \textbf{45.3}\scriptsize{$\pm$0.09} \\
        Continual (3 orders) & 2.0\scriptsize{$\pm$1.11} & 36.7\scriptsize{$\pm$2.55} & 35.1\scriptsize{$\pm$1.01} & 32.3\scriptsize{$\pm$0.88} & 18.8\scriptsize{$\pm$7.44} & \textbf{45.6}\scriptsize{$\pm$0.04} \\
    \end{tabular}%
    }
\end{table}

\textbf{More analysis on the entropy shortcut of ZO optimization.}
The logit vector $h$ can be decomposed into its scale and direction, \ie, $h=\|h\|\cdot h/\|h\|$. Our shortcut refers specifically to the scale component: the model reduces entropy simply by increasing the overall logit magnitude, without changing the relative relationships among classes, which lowers the objective but provides no meaningful task improvement. By contrast, assigning a large logit to the true class and small logits to the others alters the logit direction, and is thus a task-relevant improvement rather than a shortcut. To further verify that naive ZO entropy minimization tends to exploit this shortcut, besides the entropy and logit norm, we record two additional metrics for each sample and report their averages in Table~\ref{tab:shortcut}:
\begin{itemize}[leftmargin=12pt,itemsep=0pt,topsep=2pt]
    \item \textit{Logit-normalized entropy}: the entropy computed after rescaling logits to the average logit norm of the source model (32.44), which removes the effect of logit scale.
    \item \textit{Cosine similarity} $\cos(\hat{y}^{\mathrm{src}},\hat{y}^{\mathrm{tar}})$: the similarity between the source and adapted logits of the same input, which is scale-invariant and mainly reflects changes in logit direction.
\end{itemize}
Both BP and ZO reduce entropy while increasing the logit norm. However, after controlling for scale, BP still reduces the logit-normalized entropy from 3.33 to 2.65, indicating meaningful directional updates. In contrast, ZO only marginally reduces it to 3.20, and keeps a higher cosine similarity to the source logits (0.79 versus 0.68), \ie, it changes the logit direction less. Together with Figure~\ref{fig:motivation_1_2}~(a), these results show that naive ZO reduces entropy mainly through logit-scale inflation rather than task-relevant directional updates, which motivates our scale-invariant objective in Sec.~\ref{sec:objective_motivation}.

\begin{table}[h]
    \caption{Comparison between BP and ZO optimization under online entropy minimization on ImageNet-C (Gaussian blur, severity level 5) with ViT-Base. Logit-normalized entropy rescales logits to the average logit norm of the source model before computing entropy. $\cos(\hat{y}^{\mathrm{src}},\hat{y}^{\mathrm{tar}})$ is the cosine similarity between the source and adapted logits of the same input.}
    \vspace{-0.05in}
    \label{tab:shortcut}
    \centering
    \setlength{\tabcolsep}{7pt}
    \resizebox{0.9\linewidth}{!}{%
    \begin{tabular}{l|ccccc}
        Method & Acc. (\%) & Entropy & Logit norm & Logit-normalized entropy & $\cos(\hat{y}^{\mathrm{src}},\hat{y}^{\mathrm{tar}})$ \\
        \midrule
        NoAdapt & 35.6 & 3.39 & 32.44 & 3.33 & 1.00 \\
        BP optimization & 55.3 & 1.50 & 40.16 & 2.65 & 0.68 \\
        ZO optimization & 46.7 & 2.43 & 37.10 & 3.20 & 0.79 \\
    \end{tabular}%
    }
\end{table}

\textbf{More discussions with parameter-dynamics and anchor-based TTA methods.}
We discuss the differences between \methodname and two related BP-based TTA methods, \ie, SIGMA~\cite{lee2025parameter} and CMF~\cite{lee2024continual}.

\textit{Compared with SIGMA: different analysis target, setting, and finding.} SIGMA studies the gradient variability of effective BP learning under a Gaussian approximation of parameter transitions, comparing stable and unstable BP adaptation. In the two-forward ZO setting, however, each gradient estimate is a random direction scaled by a scalar, so its variability mainly reflects perturbation noise rather than the underlying dynamics. \methodname thus tracks the parameter deviation $\|\theta_t-\theta_0\|$ and compares ZO with BP on the same continual stream. Without the Gaussian assumption, \methodname reveals a ZO-specific phenomenon: \textit{random ZO updates accumulate substantially larger drift than BP}, which hinders long-term stability and motivates trajectory control for future ZO-based TTA.

\textit{Compared with CMF: different control mechanism--\methodname is loss-driven and more memory-efficient.} CMF maintains two extra reference states, \ie, a source model and a recursively refined source model, plus a Kalman covariance state, and fuses them with pre-scheduled weights independent of the current adaptation. In contrast, \methodname keeps only one anchor for both anchor-guided exploration and weak relaxation, reducing memory overhead. Moreover, it adapts both the anchor-relative direction and step size to the current loss difference, enabling loss-driven trajectory control that moves either toward or away from the anchor. Thus, \methodname offers a memory-efficient and loss-driven anchor-control mechanism tailored to two-forward ZO adaptation.

\subsection{Limitations, Future Directions, and Broader Impacts}\label{sec:limitations_impacts}

\methodname is mainly evaluated on standard vision tasks. Extending it to broader tasks, modalities, and real-world deployment scenarios would be an interesting direction for future work. More broadly, the two-forward, backpropagation-free nature of \methodname opens up new opportunities for efficient test-time evolution across diverse model families and deployment settings, including black-box model optimization where only forward access is available.

This work focuses on technical contributions. The proposed method is a fundamental approach that may positively support efficient and adaptive model deployment on resource-limited devices. We do not identify any direct negative societal impacts that require specific discussion.

% \section{Technical appendices and supplementary material}
% Technical appendices with additional results, figures, graphs, and proofs may be submitted with the paper submission before the full submission deadline (see above). You can upload a ZIP file for videos or code, but do not upload a separate PDF file for the appendix. There is no page limit for the technical appendices. 

% Note: Think of the appendix as ``optional reading'' for reviewers. The paper must be able to stand alone without the appendix; for example, adding critical experiments that support the main claims to an appendix is inappropriate. 

%%%%%%%%%%%%%%%%%%%%%%%%%%%%%%%%%%%%%%%%%%%%%%%%%%%%%%%%%%%%

\newpage

\end{document}